\documentclass[11pt]{article}

\usepackage[margin=1in]{geometry}

\usepackage{times}

\usepackage{setspace}
\usepackage[natbibapa]{apacite}

\usepackage{fancyhdr}
\usepackage{graphicx}
\usepackage{booktabs}
\usepackage{tabularx}
\usepackage{threeparttable}
\usepackage{amsmath}
\usepackage{subfigure}
\usepackage{subcaption}
\usepackage{siunitx}
\usepackage{enumitem}
\usepackage{algorithm}
\usepackage{algpseudocode}
\usepackage[hyphens]{url}
\usepackage{hyperref}  
\usepackage{xcolor}

\fancypagestyle{plain}{
  \fancyhf{} 
}

\fancypagestyle{titlepagestyle}{
  \fancyhf{} 
  \fancyfoot[C]{Distribution Statement A: Approved for public release; distribution is unlimited.}
}

\begin{document}

\title{Deep Learning Models for Flapping Fin Unmanned Underwater Vehicle Control System Gait Optimization

}

\author{Brian Zhou$^1$, Kamal Viswanath$^1$, Jason Geder$^1$, Alisha Sharma$^1$, Julian Lee$^{1,2}$\\$^1$: United States Naval Research Laboratory\\$^2$: Yale University}
\date{brian.zhou.ctr@nrl.navy.mil; jason.d.geder.civ@us.navy.mil; kamal.viswanath.civ@us.navy.mil; alisha.j.sharma.civ@us.navy.mil; julian.lee@yale.edu\\\textbf{Corresponding Author:} kamal.viswanath.civ@us.navy.mil}
\maketitle
\thispagestyle{titlepagestyle}

\noindent \textbf{Data Availability:} Raw data was collected at the Naval Research Laboratory. Derived data supporting the findings of this study are available from the corresponding author K.V. on request.\\
\textbf{Funding:} This work was performed at the U.S. Naval Research Laboratory through a base program funded effort.\\
\textbf{Conflict of Interest Disclosure:} The authors declare no conflicts of interest.


\newpage

\begin{abstract}
The last few decades have led to the rise of research focused on propulsion and control systems for bio-inspired unmanned underwater vehicles (UUVs), which provide more maneuverable alternatives to traditional UUVs in underwater missions. 
Recent work has explored the use of time-series neural network surrogate models to predict thrust and power from vehicle design and fin kinematics. We develop a search-based inverse model that leverages kinematics-to-thrust and kinematics-to-power neural network models for control system design. Our inverse model finds a set of fin kinematics with the multi-objective goal of reaching a target thrust under power constraints while creating a smooth kinematics transition between flapping cycles. We demonstrate how a control system integrating this inverse model can make online, cycle-to-cycle adjustments to prioritize different system objectives, with improvements in increasing thrust generation or reducing power consumption of any given movement upwards of 0.5 N and 3.0 W in a range of 2.2 N and 9.0 W.
As propulsive efficiency is of utmost importance for flapping-fin UUVs in order to extend their range and endurance for essential operations but lacks prior research, we develop a non-dimensional figure of merit (FOM), derived from measures of propulsive efficiency, that is able to evaluate different fin designs and kinematics, and allow for comparison with other bio-inspired platforms. We use the developed FOM to analyze optimal gaits and compare the performance between different fin materials, providing a better understanding of how fin materials affect thrust generation and propulsive efficiency and allowing us to inform control systems and weight for efficiency on the developed inverse gait-selector model.
\end{abstract}


\thispagestyle{empty}
\clearpage

\newpage

\setcounter{page}{1}

\section{Introduction}
Autonomous and unmanned underwater vehicles (AUV/UUVs) have a variety of industrial and research applications including exploration, mapping, and minesweeping. Historically, these operations have primarily been conducted with propeller-driven UUVs. Because propeller-driven UUVs lack maneuverability and subsequently resistance to turbulence, their operational domain is limited to relatively quiescent and deeper waters. Marine animals offer promising solutions to expanding the envelope of UUV operations because they swim with high propulsive efficiency and have high maneuverability in water \citep{masud_estimate_2022, eloy_optimal_2012, taylor_flying_2003, rohr_strouhal_2004, rohr_observations_1998, triantafyllou_hydrodynamics_2000, nedelcu_underwater_2018}, which motivates replication of their fins and other appendages in robotic designs \citep{techet_propulsive_2008}. As such, recent decades have seen the rise of research in bio-inspired propulsion systems to fill the operational gap in littoral waters and create systems with greater agility and maneuverability compared to traditional propeller-based systems. These bio-inspired propulsion systems have additional benefits: (i) multiple fins allow for better maneuvering and motion stabilization compared to conventional propulsion systems (ii) bio-inspired designs are easier to mask acoustic/hydrodynamic signatures (iii) bio-inspired fins have a lesser environmental impact compared to screw propellers or turbines which may harm vegetation and marine life \citep{Mannam2020}. Fin designs inspired by a variety of animals have been studied, including dolphins \citep{rohr_observations_1998, rohr_strouhal_2004}, penguins \citep{masud_estimate_2022}, and snakes. Among these, fish-inspired fins have driven the majority of research due to the agility these species exhibit, outperforming capabilities of traditional propulsion-based systems \citep{tangorra_development_2007, lauder_learning_2006, mignano_passing_2019, nedelcu_underwater_2018}. 

Designing, optimizing, and replicating marine flapping fin motion with bio-inspired fins requires testing of different parameters. Previous studies have extensively examined and tested the effects that different parameters such as material properties, kinematics or fin gaits, and fin shape have on a flapping fin's thrust output to better replicate and understand fish hydrodynamic performance \citep{sampath_hydrodynamics_2020, geder_maneuvering_2013, mignano_passing_2019, yun_thrust_2015, nguyen_thrust_2016, geder_underwater_nodate}. However, there is sparse literature on flapping fin control. Flapping fin UUV systems regulate the vehicle's propulsion by modifying a vehicle gait. A gait consists of a specific set of fin kinematics applied during a singular up/down-stroke cycle and is demonstrated in Table \ref{tab:param}. 

While previous research has studied the effect of various kinematics on propulsion through experimental, computational fluid dynamics, and surrogate model techniques, all three approaches currently fail to embed a full understanding of how a complete set of kinematic aspects) can affect the propulsion and power consumption of any specific gait. For example, existing approaches either focus on experimentally determining a small set of high-propulsion gaits \citep{88fb8aabc2e0433bb6a8c074c31578f4}, restrict chosen gaits to a line in the multi-dimensional kinematic space \citep{bietal14}, or incrementally changes kinematics that correlates positively or negatively with thrust to gradually approach the optimal target propulsion \citep{Palmisano2008}. Every current approach to find an optimal gait is limited to a small set of the millions of possible kinematics and is bound to miss on the highest-propulsion or most efficient gait possible to achieve a desired movement. Additionally, a previously proposed inverse control model is limited in only studying various subsets of the drag-force generation; this prevents the proposed model from being power-aware and optimizing for efficiency and power-related objectives \citep{Remmas2021}. We use a dual neural network model to embed a more comprehensive understanding of the relationship between gait and propulsion as well as gait and power consumption by predicting the power consumed and thrust generated from the time series of any given gait; as a result, the control system can not only generate gaits that meet a desired trajectory, but also optimize for important performance measures such as rapid propulsion acceleration, a smooth motor transition, and a greater energy efficiency.

\begin{figure}[h!]
\centerline{\includegraphics[width=1\linewidth,height=\textheight,keepaspectratio]{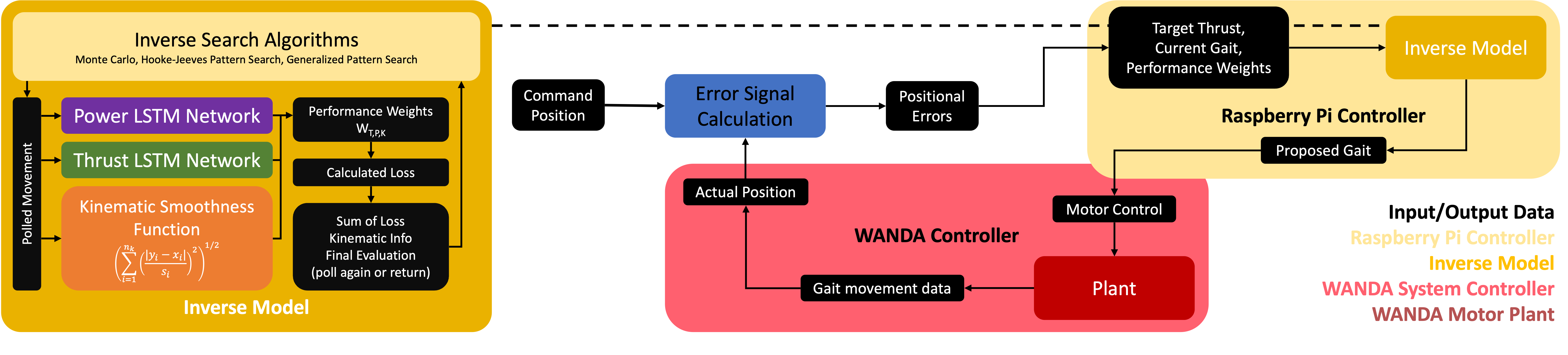}}
\caption{Integration of the inverse model (gold) within the control system.}
\label{fig:inversesystem}
\end{figure}

We propose a novel approach to gait generation that uses a search-based inverse model to prompt two forward surrogate models. The UUV control system provides a requested destination or speed, which will be broken down into individual cycle-by-cycle thrust requests which are mapped to a gait to achieve the desired multi-step goal. The inverse model uses set performance metric weights (optimized for propulsion, efficiency, motor smoothness, etc.) to search through the space of all possible gaits to find and return the best subsequent gait that both meets the thrust requirements and optimizes for other values. Other approaches, such as direct neural networks \citep{Remmas2021, cali_adaptive_2020, muliadi_neural_2018}, are unable to create this flexible optimizer able to change priorities cycle-by-cycle to prioritize different metrics. To evaluate the thrust generated or power consumed by any specific gait, a neural network forward model predicts the thrust and power consumed of any given gait. Figure \ref{fig:inversesystem} shows the integration of the inverse model and neural networks within the control system.

In this paper, we develop and test multiple forward gait-to-thrust and gait-to-power models, demonstrating that models are able to accurately interpret gaits and comparing different deep learning techniques to improve model accuracy. While prior studies have emphasized solely studying the effects of designs and gaits on propulsion \citep{sampath_hydrodynamics_2020, geder_maneuvering_2013, mignano_passing_2019, yun_thrust_2015, nguyen_thrust_2016, geder_underwater_nodate}, we resolve a gap in literature with the effects that designs and gaits have on power consumption and UUV efficiency; previous research in power utilizes CFD simulations to study hydrodynamic power ineffective for use in a control system that requires accounting for power loss of integrated actuators \citep{palmisano_power_nodate}. To offer comparability for other UUV systems in the future, we develop a dimensionless figure of merit (FOM) that can compare the efficiency between different systems.

We show that the  thrust-to-position search-based inverse model is able to embed performance weights to find efficient or high propulsion gaits and compare between different search techniques to improve inverse model performance. A challenge for most deep-learning approaches is that bio-inspired UUVs are typically small and lightweight to maximize the advantages compared to conventional propulsion systems, limiting battery size and electronic compute power. Through benchmarking on a Raspberry Pi, We demonstrate that my inverse search model fulfills computational and time constraints, even with the use of an expensive time-series forward model. 

We test performance gains provided from the inverse model through model simulations via MATLAB's Simulink environment and physical platform testing. Using test results, we construct a series of optimal weight combinations tailored for various operational requirements such as power-saving, rapid acceleration, and station keeping modes by evaluating the trade-offs of propulsion and efficiency in various gaits. Beyond the strength of the inverse search model in live cycle-by-cycle performance improvements, We demonstrate the strength of the forward gait-to-thrust/power and inverse position-to-gait model as a framework to evaluate the propulsion and efficiency of three potential fin material designs.

\section{Materials and Methods}
\subsection{Experimental Setup}
\begin{figure}[h!]
\begin{minipage}[c]{0.45\linewidth}
\includegraphics[width=\linewidth]{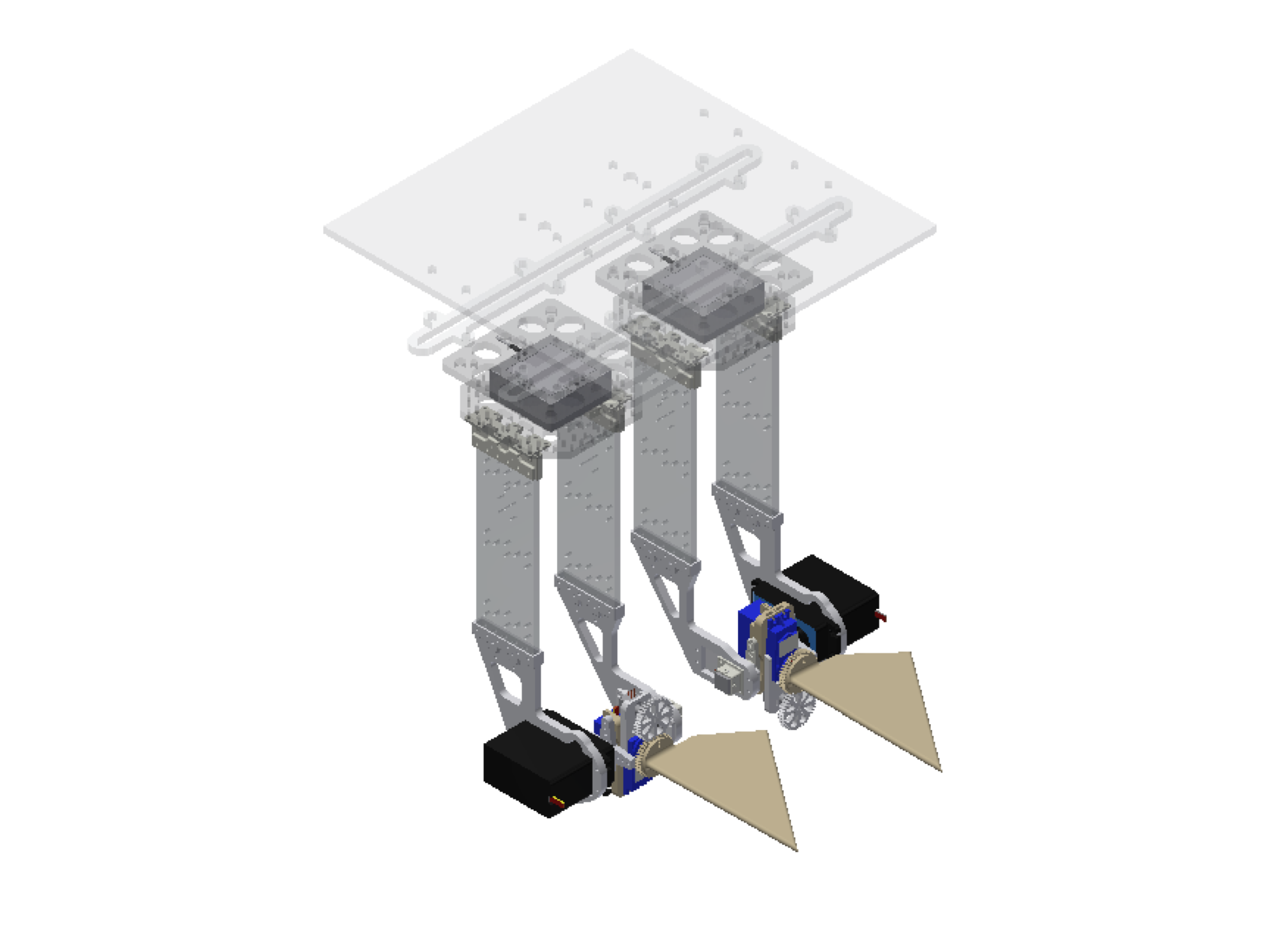}
\caption{CAD design of tandem fins mounted to instrumentation and control platform.}
\label{fig:cad-multifin}
\end{minipage}
\hfill
\begin{minipage}[c]{0.45\linewidth}
\includegraphics[width=\linewidth]{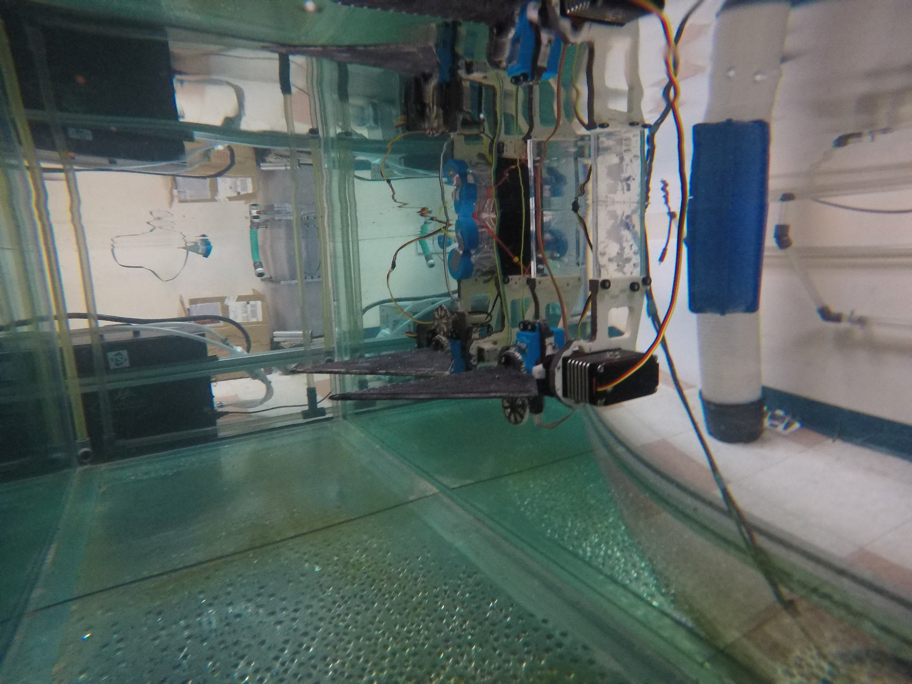}
\caption{Tandem fin testing platform in experimental test environment.}
\label{fig:tank}
\end{minipage}%
\end{figure}

To train and evaluate both the forward and inverse models, experimental data was collected for a system of artificial pectoral fins mounted in an underwater tank shown in Figures \ref{fig:cad-multifin} and \ref{fig:tank}. As prior research has demonstrated that tandem fin configurations have minimal effects on thrust output and power consumption,
we only run single-fin tests to gather data. 



The control platform was mounted in a 2.41 x 0.76 x 0.76 m (length, width, height) glass tank. A microcontroller controlled the fin actuators to collect data on the programmed gait combinations as laid out in Tables \ref{tab:param} and \ref{tab:combo}. Potentiometers (TT Electronics P260) measure the stroke and pitch angles over time, while load cells (Interface 3A60A) measure the generated forces \citep{geder_fluid-structure_2021}.

\subsection{Parameters and Outputs}
\begin{figure}[h!]
\centerline{\includegraphics[width=0.55\linewidth,height=\textheight,keepaspectratio]{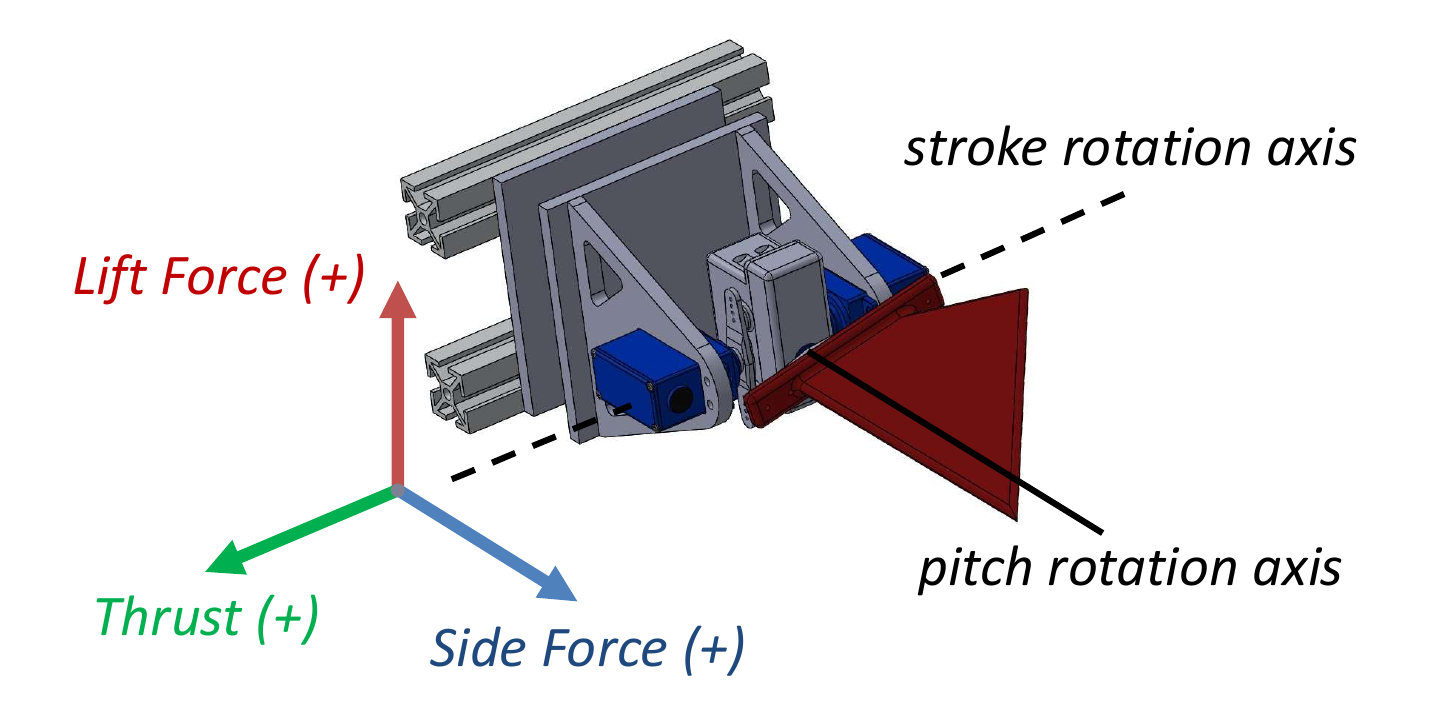}}
\caption{CAD design of a single flapping fin propulsor with coordinate reference frames.}
\label{fig:coords}
\end{figure}

A singular gait is comprised of four kinematic parameters: stroke phase offset, stroke amplitude, pitch amplitude, and flapping frequency (Table \ref{tab:param}). For each unique gait, we collected experimental data on the forces generated in the x/y/z axis, current drawn by each motor, and circuit voltage for the different fin gaits, which we use to develop the forward models for thrust force and power. Although only the static kinematic values create a unique gait, dynamic kinematic values are still helpful for more accurate model training and interpolation. This information is laid out in Table \ref{tab:out}, and force vectors are defined in Figure \ref{fig:coords}. 

\begin{table}[ht]
  \caption{Kinematic Gait Parameters}
  \label{tab:param}
  \centering 
  \begin{threeparttable}
    \begin{tabular}{>{\raggedright\arraybackslash}p{0.25\linewidth} p{0.1\linewidth} p{0.55\linewidth}}
     \midrule\midrule
    \textbf{Parameter} & \textbf{Symbol} & \textbf{Description} \\
     \midrule\midrule
    \multicolumn{3}{c}{\textit{Static Kinematics}} \\
    \cmidrule(l r ){1-3}
    Frequency (Hz) & $f$ & Number of flap cycles per second \\
    \cmidrule(l r ){1-3}
    Stroke Pitch Offset (\si{\degree}) & $\delta$ & Phase offset of the pitch cycle to the stroke cycle, calculated as $\frac{1}{16}$th of one cycle \\
    \cmidrule(l r ){1-3}
    Stroke Amplitude (\si{\degree}) & $\Phi$ & Maximum stroke angle over one flap cycle \\
    \cmidrule(l r ){1-3}
    Pitch Amplitude (\si{\degree}) & $\Theta$ & Maximum pitch angle over one flap cycle \\
    \cmidrule(l r ){1-3}
    \multicolumn{3}{c}{\textit{Dynamic Kinematics}} \\
    \cmidrule(l r ){1-3}
    Stroke Angle (\si{\degree}) & $\phi$ & Time history of stroke angle \\
    \cmidrule(l r ){1-3}
    Pitch Angle (\si{\degree}) & $\theta$ & Time history of pitch angle \\
    \midrule\midrule
    \end{tabular}
\end{threeparttable}
  \end{table}
 
\begin{table}[h!]
  \caption{Control System Measurements}
  \label{tab:out}
  \centering 
  \begin{threeparttable}
    \begin{tabular}{>{\raggedright\arraybackslash}p{0.25\linewidth} p{0.1\linewidth} p{0.55\linewidth}}
     \midrule
     \midrule
    \textbf{Parameter} & \textbf{Symbol} & \textbf{Description} \\
     \midrule
     \midrule
    \multicolumn{3}{c}{\textit{XYZ Forces}} \\
    \cmidrule(l r ){1-3}
    Thrust (N) & $T$ & Force generated along stroke axis \\
    \cmidrule(l r ){1-3}
    Lift (N) & $L$ & Force generated perpendicular to both \\
    \cmidrule(l r ){1-3}
    Side Force (N) & $S$ & Force generated along pitch axis \\
    \cmidrule(l r ){1-3}
    \multicolumn{3}{c}{\textit{Power Consumption}} \\
    \cmidrule(l r ){1-3}
    Stroke Current (A) & $I_{\phi}$ & Time history of current draw for the stroke actuator \\
    \cmidrule(l r ){1-3}
    Pitch Current (A) & $I_{\theta}$ & Time history of current draw for the pitch actuator \\
    \cmidrule(l r ){1-3}
    Voltage ($V$) & $V$ & Voltage of both actuators \\
    \midrule\midrule
    \end{tabular}
\end{threeparttable}
  \end{table}

\subsection{Materials}
While previous studies have tested various properties of the fin design including fin shape, fins material stiffness \citep{geder_fluid-structure_2021} and fin configuration \citep{sampath_hydrodynamics_2020,geder_effect_2018}, we introduce a single fin design parameter: material flexibility. Three different fin designs in total are introduced in Table \ref{tab:mat}.

\begin{table}[h!]
  \caption{Fin Material Properties}
  \label{tab:mat}
  \centering 
  \begin{threeparttable}
    \begin{tabular}{p{0.4\linewidth} p{0.2\linewidth}}
     \midrule\midrule
    \textbf{Material} & \textbf{Young's Modulus}  \\
     \midrule\midrule
    Rigid Nylon & ~1 GPa \\
    \cmidrule(l r ){1-2}
    Polydimethylsiloxane (PDMS) 1:10 & 850 kPa \\
    \cmidrule(l r ){1-2}
    Polydimethylsiloxane (PDMS) 1:20 & 310 kPa \\
    \midrule\midrule
    \end{tabular}
\end{threeparttable}
  \end{table}

\subsection{Data Collection} 
\begin{figure}[h!]
\centerline{\includegraphics[width=0.8\linewidth,height=\textheight,keepaspectratio]{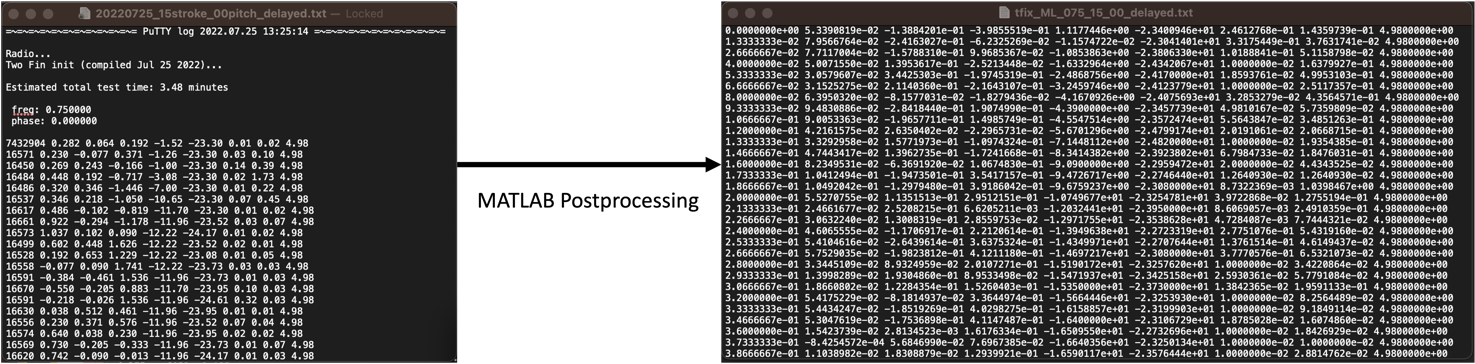}}
\caption{An example post-processing sequence of a control system sensor output log to training data.}
\label{fig:postprocessing}
\end{figure}

The parameter space for the kinematics variables is large. Experimental data is first constrained to parameter values that are physically achievable given an operating frequency. At higher frequencies, the fins are physically unable to reach certain stroke and pitch amplitudes, as the duration of the gait is shorter. Equation \ref{eq:1} defines the range of achievable strokes and pitches with respect to frequency: 
\begin{equation}
    \label{eq:1}
    \text{Attainable gaits:} \begin{cases}
        \text{$0<\Phi<97-f*30$}\\
        \text{$0<\Theta<75-f*26$}\\
    \end{cases}  
\end{equation}
With this equation, we collected data that covers the entire scope of the achievable kinematics range at each frequency. In total, 864 unique gait combinations listed in Table \ref{tab:combo} were tested, as it covered the full range of gaits while taking a reasonable amount of time to collect data. Each cycle takes approximately 2 minutes, making the total time for all three materials upwards of 80 hours. For each fin gait, ten flap cycles were run. Only the five middle cycles were used for analysis to account for discrepancies when the actuator started and ended the cycle motions. Recorded sensor data was converted into final values using a MATLAB post-processing script, as shown in Figure \ref{fig:postprocessing}. Data was collected in a zero velocity flow condition, which previous research has demonstrated is important for low-speed maneuvering to station-keep and offset buoyancy \citep{geder_fluid-structure_2021}. 

This process was replicated for all three fin material designs. Taking the experimental data as defined in Table \ref{tab:out}, we computed the total power consumption of both the stroke and pitch actuators with Equation \ref{eq:2}.
\begin{equation}
    \label{eq:2}
    P = I_{\phi} * V + I_{\theta} * V
\end{equation}
  
\begin{table}[ht]
  \caption{Gait Combinations}
  \label{tab:combo}
  \centering 
  \begin{threeparttable}
    \begin{tabular}{>{\raggedright\arraybackslash}p{0.3\linewidth} p{0.3\linewidth}}
     \midrule
     \midrule
    \textbf{Parameter} & \textbf{Values}  \\
     \midrule
     \midrule
    Stroke Amplitude (\si{\degree}) & $0, 15, 25, 32.5, 40, 55$  \\
    \cmidrule(l r ){1-2}
    Pitch Amplitude (\si{\degree}) & $0, 15, 25, 32, 38, 55$ \\
    \cmidrule(l r ){1-2}
    Frequency (Hz) & $0.75, 1.00, 1.25, 1.50, 1.75, 2.00$ \\
    \cmidrule(l r ){1-2}
    Stroke Pitch Offset (\si{\degree}) & $-22.5, 0, 22.5, 45$  \\
    \cmidrule(l r ){1-2}
    Voltage ($V$) & Constant at $4.98V$ \\
    \midrule\midrule
    \end{tabular}
\end{threeparttable}
  \end{table}

\section{The Forward Model}
Two forward models predict either the average UUV thrust or the average power consumed for a flapping cycle from the static gait information given. While reduced-order analytic models can produce fast predictions, they struggle to maintain accuracy when generalized beyond a small parameter space \citep{97c98750ccb7468ebb8bc45454d70003}. A model supporting a higher-order input space will allow future forward gait-to-propulsion models to incorporate fluid dynamics-related parameters such as flow speed as well as multi-fin kinematic parameters such as the flapping phase offset between front and rear fins.

Neural network surrogate models support higher degree input spaces, and prior flapping fin propulsion research on fin design has developed neural network surrogate models for thrust prediction \citep{viswanath2019evaluation,Lee2021ThrustPrediction}. Both works demonstrate that time series models can predict the time history of thrust generation for a flapping cycle. Therefore, we implement a similar long-term short memory (LSTM) network with the gait-to-thrust forward model from \cite{lee_data-driven_2023} using the inputs in Table \ref{tab:param} and compare the accuracy and runtime performance with other reduced-order and high-order approaches. LSTM networks process sequential data by generating an output at each time step and using information from past outputs to inform subsequent results. Compared to traditional recurrent neural networks, LSTM networks include a cell state to retain a long-term memory accumulated from multiple past time steps that influences the next output.

\subsection{Objectives}
Three objectives guided the selection and development of a high-order model:
\begin{enumerate}[noitemsep]
    \item Completion of a baseline model that can accurately take in kinematics data (frequency, stroke amplitude, pitch amplitude, and offset) to output predicted power
    \item Capability to take in static information such as material and flexibility to use different models to maximize accuracy and usefulness of integrated model
    \item Run-time speed of 100 forward passes per second at minimum, with computational power not exceeding the capacity of a Raspberry Pi
\end{enumerate}

Separate forward models are developed for various design-related parameters such as material, design, rigidity, and tandem fin spacing. Since these parameters installed are static and will not be replaced during missions, a forward model is more accurate for cycle-by-cycle calculations and only needs to load what is relevant to the mission without wasting excess computational power. Forward models are trained on dynamic gait information such as frequency, stroke angle, pitch angle, angle offset, and tandem phasing.

\subsection{Model Results}
We examined six approaches to model thrust and power; two were reduced-order polynomial models to serve as a baseline and four utilized high-order deep learning approaches. 

\begin{figure}[h!]
\centerline{\includegraphics[width=0.6\linewidth,height=\textheight,keepaspectratio]{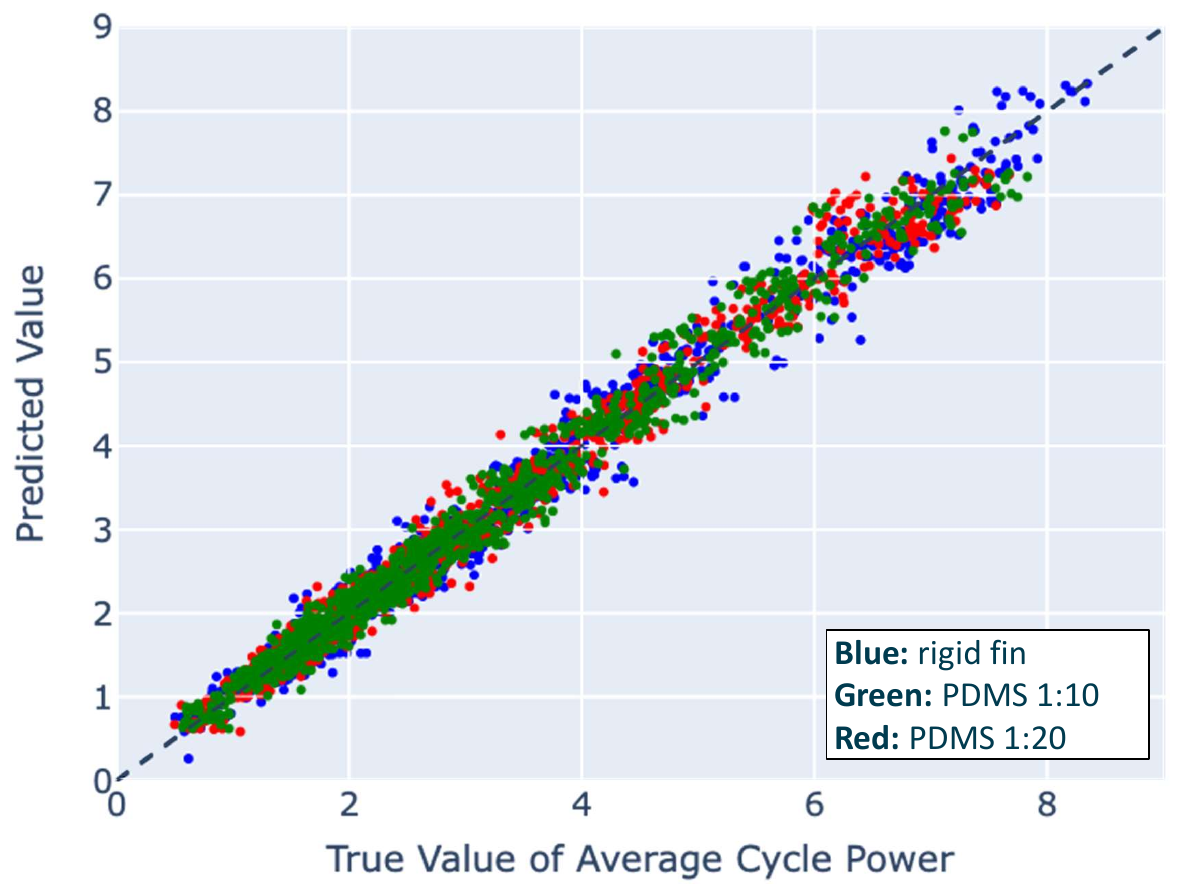}}
\caption{Reduced-order quartic polynomial model performance for synthetic data, comparing predicted power values to the true power value. Colors indicate the material, with blue (rigid) green (1:10) red (1:20)}
\label{fig:model}
\end{figure}

We explore linear models from degrees one to five and found that the quartic model best fit the data. Using the linear model as a baseline, the model performed the fastest with an average error across all 3 fin designs of 0.3891 W and 0.1638 N. As the true value and predicted values appeared to have an exponential relationship in the linear model, we tested a quartic polynomial which produced better results (Figure \ref{fig:model}) with an average error of 0.1815 W and 0.0911 N. Out of the three data sets, the PDMS 1:10 fins fit the best to the quartic regression.

To explore high-order approaches, we implement a Multi Layer Perceptron Regression (MLP), a Convolutional Neural Network (CNN), a Dense Neural Network (DNN), and a LSTM model that uses the static kinematic values in Table \ref{tab:param} to directly predict average thrust and power. Of these high-order approaches that do not utilize time series, DNNs, which consist of layers of nodes such that each node in layer $l$ is connected to every node in layer $l-1$, perform the best. Overall, deep learning approaches increase the accuracy while operating at a speed fast enough for model integration. Table \ref{tab:model} highlights the benchmarked results and relative time performances for various forward model approaches. Power ranged from 0 to 9.3 W, and propulsion ranged from -1.1 to 1.1 N. Each model is benchmarked on a Raspberry Pi Model 4B, and although the LSTM is approaches the limit, it is able to make 100 computations/second Pi.

\begin{table}[ht]
  \caption{Model Performances}
  \label{tab:model}
  \centering 
  \begin{threeparttable}
    \begin{tabular}{>{\raggedright\arraybackslash}p{0.25\linewidth} p{0.25\linewidth}  p{0.25\linewidth} p{0.1\linewidth}}
     \midrule
     \midrule
    \textbf{Model} & \textbf{Averaged Error (W)} & \textbf{Averaged Error (N)} & \textbf{Runtime}  \\
     \midrule
     \midrule
    Linear Polynomial & 0.3891 & 0.1638 & Low  \\
    \cmidrule(l r ){1-4}
    Quartic Polynomial & 0.1815 & 0.0911 & Low \\
    \cmidrule(l r ){1-4}
    MLP & 0.1229 & 0.0638& Medium \\
    \cmidrule(l r ){1-4}
    CNN & 0.0907 & 0.0572& Medium \\
    \cmidrule(l r ){1-4}
    DNN & 0.0429& 0.0364 & Medium \\
    \cmidrule(l r ){1-4}
    LSTM & 0.0072 & 0.0076& High \\
    \midrule
    \midrule
    \end{tabular}
\end{threeparttable}
  \end{table}

Out of all five power models, the LSTM has the best performance but is also the most time-consuming. An additional benefit of the LSTM is that it can utilize the dynamic time series inputs to accurately interpolate between gaps of data, which is useful for understanding how the whole gait space behaves for various movements.

When trained on all experimental gaits, the thrust LSTM reached an average error of 0.0076 N and the power LSTM reached an average error of 0.0072 W. The most visible shortfall of the power LSTM model is its inability to grasp the time history of power consumption or thrust at certain gaits. Each cycle is vastly different at a low flapping fin frequency combined with a low stroke angle due to random actuator noise creating jolts in the time history that would otherwise be negligible in higher frequency gaits. Even still, it is able to predict the average time history of many cycles, and for the purpose of interpolating the final numerical power consumption or propulsion generated of any given gait, this is not a setback.

\subsection{LSTM Results}
\begin{figure}[h!]
\centerline{\includegraphics[width=\linewidth,height=\textheight,keepaspectratio]{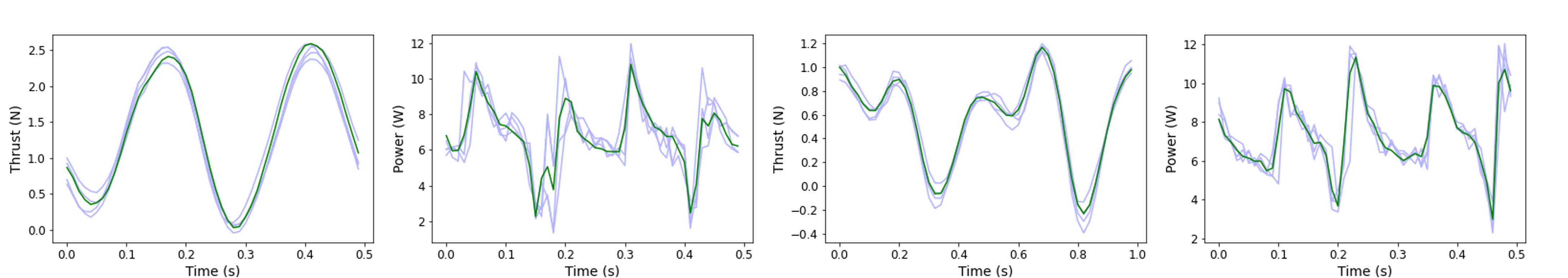}}
\caption{Two sample experimental (blue) and predicted (green)
thrust/power time histories for two interpolated kinematics. The left
graph involves interpolation to an unseen stroke and pitch
angle, while the right graph involves interpolation to an unseen flap frequency and stroke pitch offset}
\label{fig:samplegaits}
\end{figure}

We train each LSTM model to 1000 epochs for modeling both power and thrust predictions. The statistics for the LSTM models produced are found in Table \ref{tab:LSTM}. 

\begin{table}[ht]
  \caption{Average Error for LSTM Models}
  \label{tab:LSTM}
  \centering 
  \begin{threeparttable}
    \begin{tabular}{p{0.3\linewidth} p{0.15\linewidth} p{0.15\linewidth} p{0.15\linewidth}}
     \midrule
     \midrule
    \textbf{Interpolations} & \textbf{Rigid} & \textbf{PDMS 1:10} & \textbf{PDMS 1:20}  \\
     \midrule
     \midrule
    Power (W) & 0.0236 & 0.072 & 0.0268 \\
    \cmidrule(l r ){1-4}
    Thrust (N) & 0.0002 & 0.0186 & 0.0041 \\
    \midrule
    \midrule
    \end{tabular}
\end{threeparttable}
  \end{table}

To test LSTM gait interpolation, a holdout set of gaits was excluded from training. My holdout set consisted of all experimental gaits fulfilling one or more of the following criteria: a flap frequency of 1.25 Hz, a stroke-pitch offset of 0, or a stroke or pitch amplitude of 25\si{\degree}. The LSTM successfully interpolated kinematics for the excluded gaits with a mean average thrust error of 0.0344N. The worst performing subset of excluded gaits–gaits with a stroke pitch offset of 0–still obtained a mean average thrust error of 0.0374N. Figure \ref{fig:samplegaits} shows example thrust time histories generated by the LSTM for interpolated gaits. The LSTM embeds an understanding of how thrust changes over the course of a flapping cycle, capturing the peak and troughs of the thrust time history; this understanding offers an explanation for the high-accuracy LSTM average thrust predictions for interpolated kinematics.

\begin{figure}
\begin{minipage}[c]{0.3\linewidth}
\includegraphics[width=\linewidth]{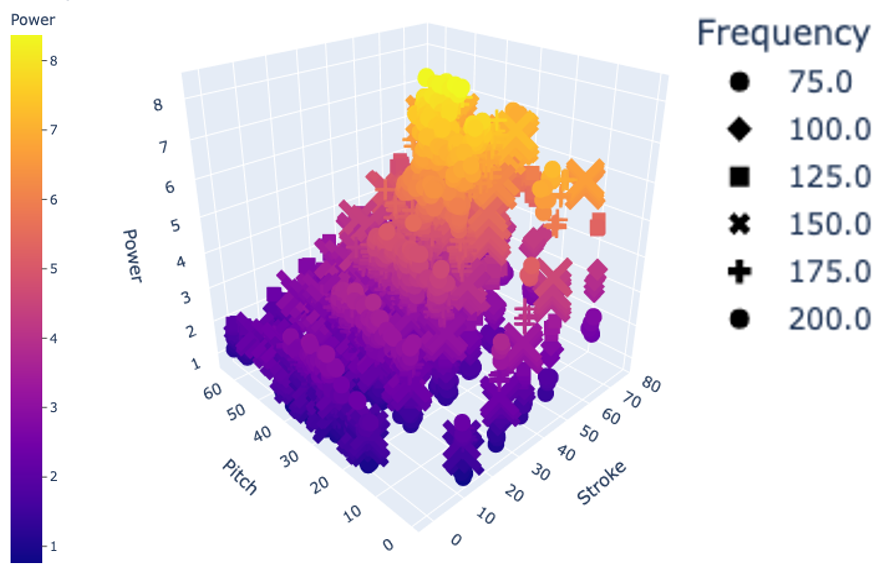}
\caption{Space of 864 gaits.}
\label{fig:origdata}
\end{minipage}
\hfill
\begin{minipage}[c]{0.3\linewidth}
\includegraphics[width=\linewidth]{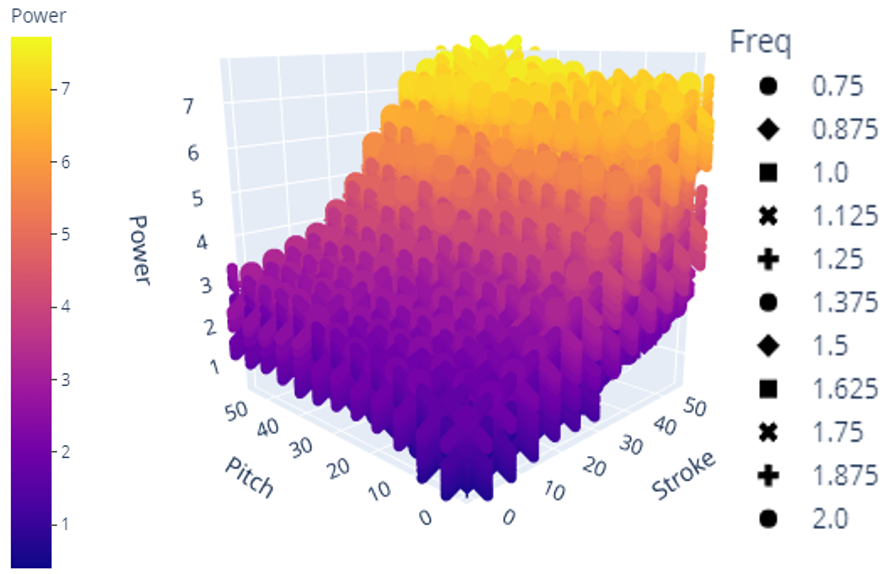}
\caption{Space of 20,591 gaits.}
\label{fig:20kdata}
\end{minipage}
\hfill
\begin{minipage}[c]{0.3\linewidth}
\includegraphics[width=\linewidth]{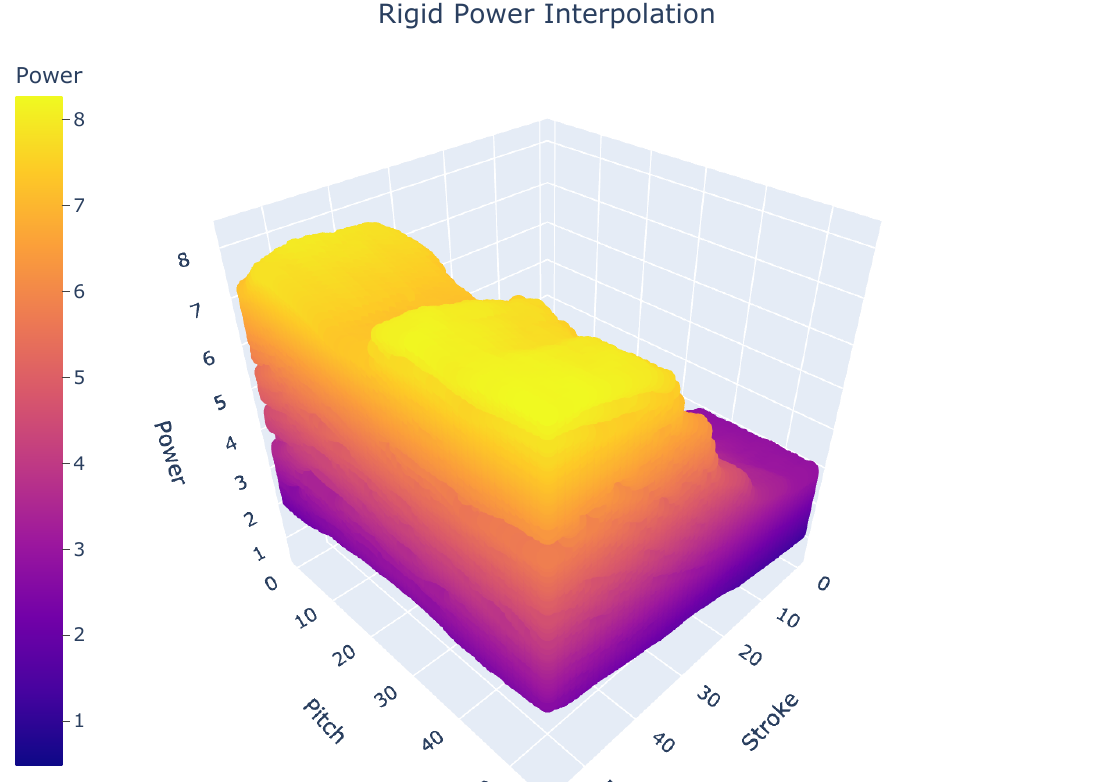}
\caption{Space of 400k+ gaits.}
\label{fig:gendata}
\end{minipage}%
\end{figure}

By testing the ability of an LSTM to interpolate between large holdout sets in data, we also demonstrate that an LSTM is effective at filling between intentional gaps in data. For example, a LSTM is able to turn the dataset of 864 unique gait combinations visualized in Figure \ref{fig:origdata} to any desired level of depth, including 20,591 gaits in Figure \ref{fig:20kdata} and more than 400,000 gaits in Figure \ref{fig:gendata}.

In total, we generated interpolations within the constraints given by the collected data outlined in Equation \ref{eq:1}. We interpolated data for every stroke and pitch combination from 0 to 55 degrees with 1 degree increments, frequency from 0.75 to 2 Hz with 0.125 Hz increments, and SPO from -22.5 to 45 degrees with 5.625 degree increments. In total, 435,600 interpolations were calculated for each data set. A sample of the rigid fin data space is shown in Figure \ref{fig:gendata}. The interpolations filled gaps of data, creating better insight into how different gait patterns and materials behave for various of-interest combinations such as at very low frequencies and flapping angles or high frequencies and high angle offsets.

\section{Figure of Merit}
\subsection{Objectives}
To measure propulsive efficiency of a vehicle or thruster, traditionally thrust is multiplied by vehicle velocity and divided by power input. Because our thrust results were achieved at zero freestream flow to understand force production near hover and low speed operations, we create a dimensionless figure of merit (FOM) as a surrogate for propulsive efficiency.

\subsection{Development}
To compute the FOM $\eta$, the average thrust over a cycle is multiplied by a characteristic velocity scale and divided by the flapping cycle power input, as displayed in Equation \ref{eq:fom1}. This non-dimensional value allows us to compare across different gaits and inflow conditions. For vehicle integration purposes, this allows for the return of a pure value or a percentage compared to the highest recorded gait FOM for a loaded design.

\begin{equation}
    \label{eq:fom1}
    \eta  = \frac{F_{avg}*v}{P_{avg}}
\end{equation}


Since all tests were conducted in a constant zero flow, the velocity term is initially set to 1 m/s for relative comparisons between tests. However, to allow for better comparison across different stroke frequencies and amplitudes in later tests, a universal FOM utilizes the average fin tip speed as the characteristic velocity, as shown in Equation \ref{eq:fom4}, where $\Phi$ is the stroke amplitude, $r_{tip}$ is the distance from the rotation axis to the fin tip, and $f$ is the flapping frequency. 
\begin{equation}
    \label{eq:fom4}
    v  = 2\pi*(\frac{4*\Phi}{360})*r_{tip}*f
\end{equation}
 

A FOM can also adapt to different objectives. While we study the specific propulsive efficiency of thrust, the FOM can isolate how efficiency changes with the individual stroke and pitch actuators or examine the side and lift forces.




\subsection{FOM Results}
\begin{figure*}[p]
\centering
\includegraphics[width=\linewidth]{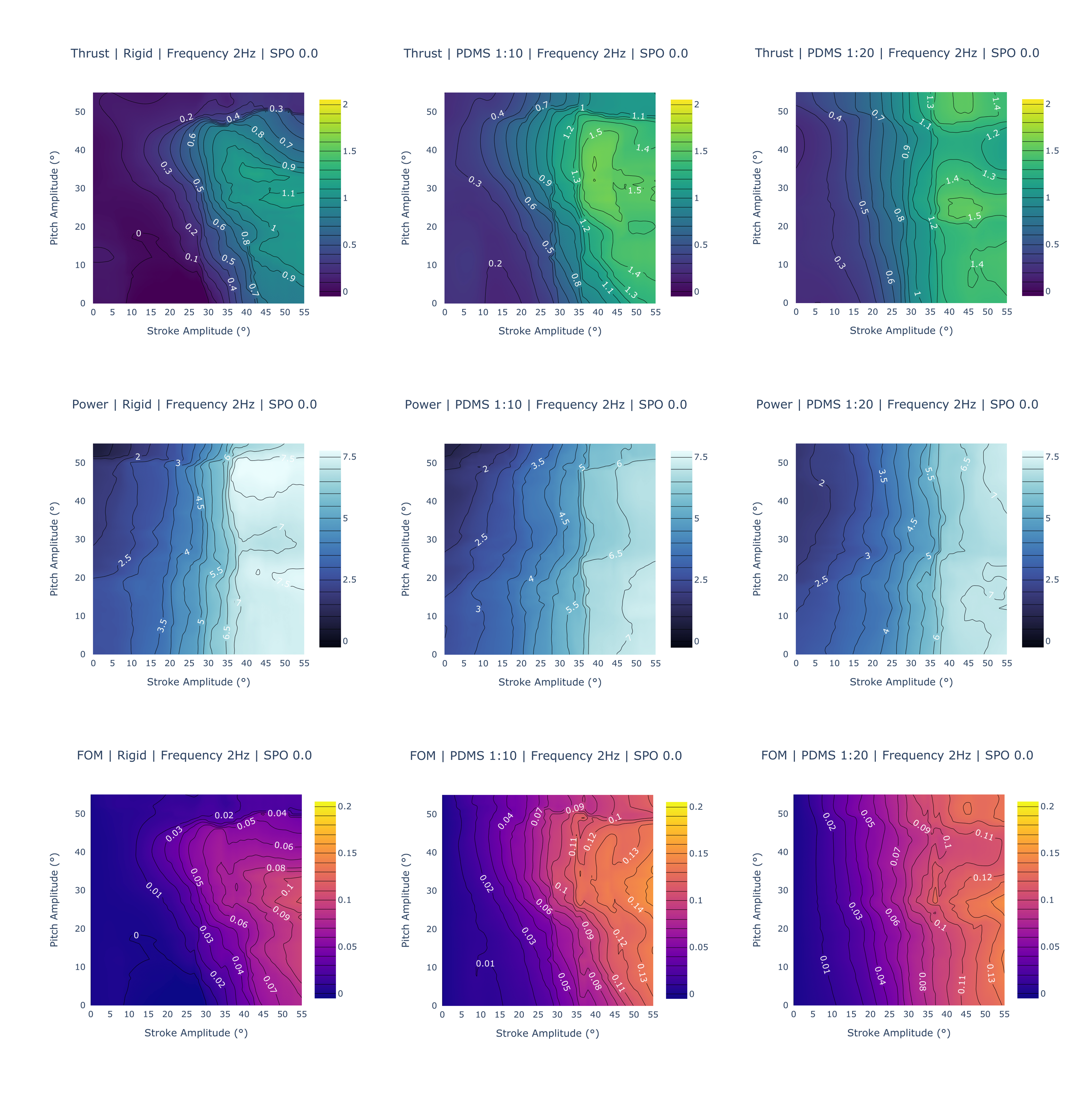}
\caption{Thrust, power, and figure of merit contours. The columns are different data sets (rigid, PDMS 1:10, and PDMS 1:20) while the rows graph different gait results (Thrust, Power, and FOM value). The PDMS 1:10 fin generates the highest possible thrust and is higher overall in more gaits; the rigid fin is significantly worse at thrust generation than either of the PDMS fins. The PDMS 1:10 and 1:20 fins are comparable in power consumption but differ in trends at higher stroke and pitch combinations; the rigid fin consumes significantly more power. The PDMS 1:10 fin has the largest FOM values, with the PDMS 1:20 fin following. The rigid fin is significantly worse in all 3 metrics.}
\label{fig:fom}
\end{figure*}

\begin{figure*}[t!]
\centering
\includegraphics[width=\textwidth,height=\textheight,keepaspectratio]{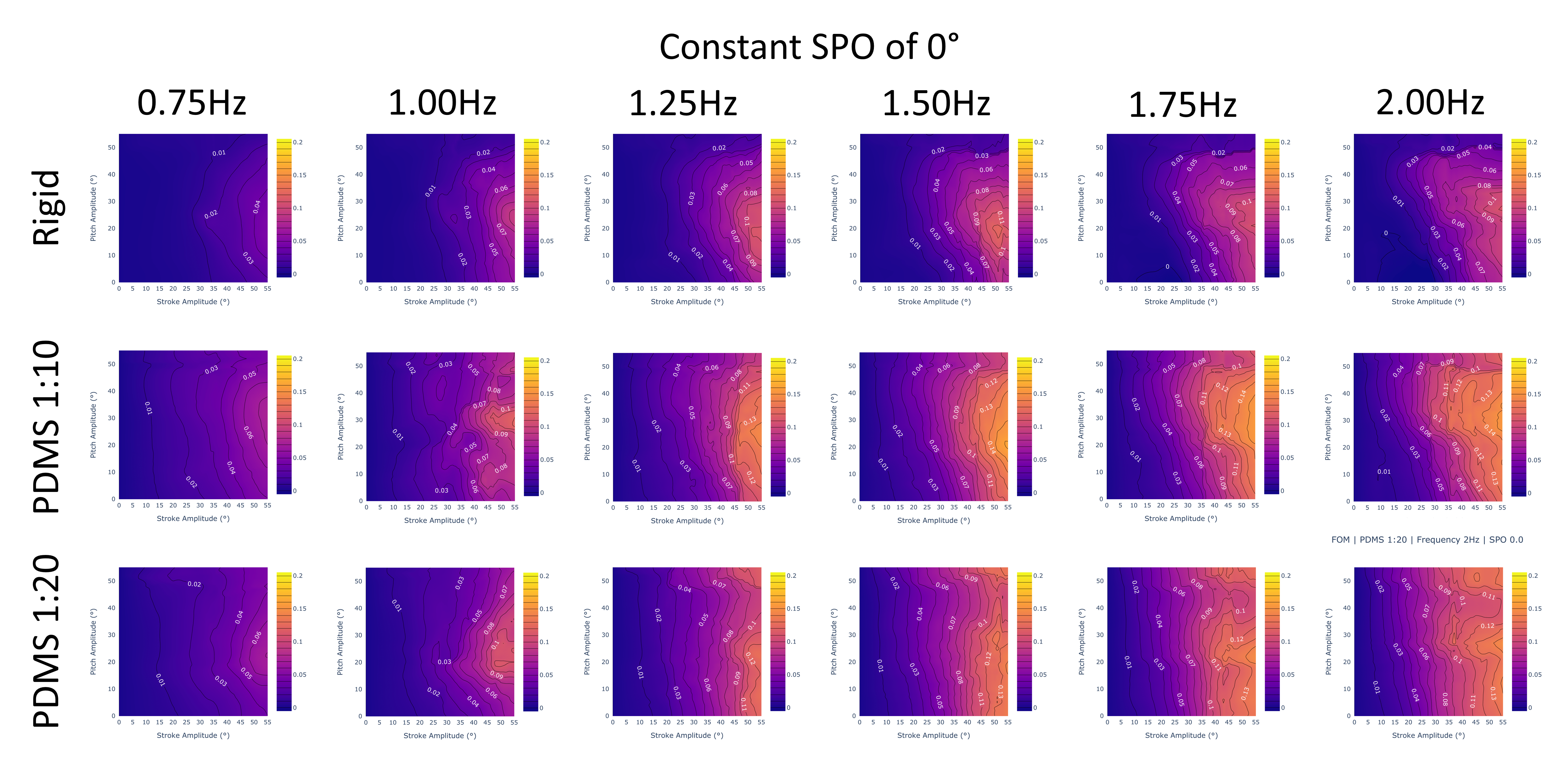}
\caption{Figure of Merit results for a constant Stroke-Pitch-Offset of 0 degrees and varying frequency. The rows graph all interpolated frequency values with 0.25Hz increments while the columns are different data sets (rigid, PDMS 1:10, and PDMS 1:20). we verify that PDMS 1:10 is the best performing across all frequencies with higher FOM values and averages compared to the PDMS 1:20 or rigid fins. Additionally, increasing the frequency will improve the FOM regardless of fin design.}
\label{fig:freq}
\end{figure*}

To gain a better understanding of how the figure of merit correlates to the thrust force and power consumption, we created a grid of contours for one gait, shown in Figure \ref{fig:fom}. Here, the stroke and pitch range for a frequency of 2 Hz and stroke-pitch offset of 0$^\circ$ is depicted. The grid contains columns with the three material data sets and each row graphs part of the figure of merit equation: thrust, power, and FOM in that order. 

Beginning with the thrust, a few trends are immediately apparent. First, the fin design that can generate the largest force is the PDMS 1:10 design, generating a maximum force of around 1.6 N at a 40$^\circ$ stroke amplitude and 30-40$^\circ$ pitch amplitude. Both the other designs trail behind, with the PDMS 1:20 fin being able to generate a thrust of 1.5 N and the rigid fin only managing up to 1.1 N. The PDMS 1:10 fin also has more gaits at higher thrust levels. An observation of the contour reveals that there are more combinations of stroke and pitch that produce higher thrusts when compared to the PDMS 1:20 fin. An analysis of all gaits including other variable stroke-pitch offsets and frequencies confirm these findings. The PDMS 1:10 fin's maximum thrust is 2.1 N, while the PDMS 1:20 fin produces a maximum force of 1.6 N and the rigid fin produces a maximum force of 1.2 N. The PDMS 1:10 fin has the highest average thrust generation, followed by the PDMS 1:20 fin.

Power consumption goes in the reverse order. The design that requires the highest wattage is the rigid fin design, requiring 7.6 W for any gait with a stroke amplitude of more than 40$^\circ$. The PDMS 1:10 and PDMS 1:20 fins are similar, with a maximum power consumption of 7.1 W. However, at cases above 40$^\circ$ stroke and 30$^\circ$ pitch, the PDMS 1:10 fin observes lower wattage consumed at the same gait combination. The PDMS 1:20 fin appears to depend less on the pitch amplitude, with power more dependent on the stroke amplitude. These observations are verified when looking at all gaits. While the PDMS fins have a maximum wattage of around 7.5 W, it occurs at much fewer gaits than with the rigid data set. 

Another interesting trend is visible when comparing the FOM and thrust charts, which appear almost identical with only a few differences in their trends. The explanation becomes evident when looking at the contours for power, which have a near-linear trend across stroke amplitude. While pitch amplitude does affect both the PDMS 1:10 and 1:20 designs, the stroke amplitude has the most recognizable and significant effect. Future work will include generating additional figures to verify that this trend exists across all frequencies and stroke-pitch offsets.

\begin{figure*}[t!]
\centering
\includegraphics[width=\textwidth,height=\textheight,keepaspectratio]{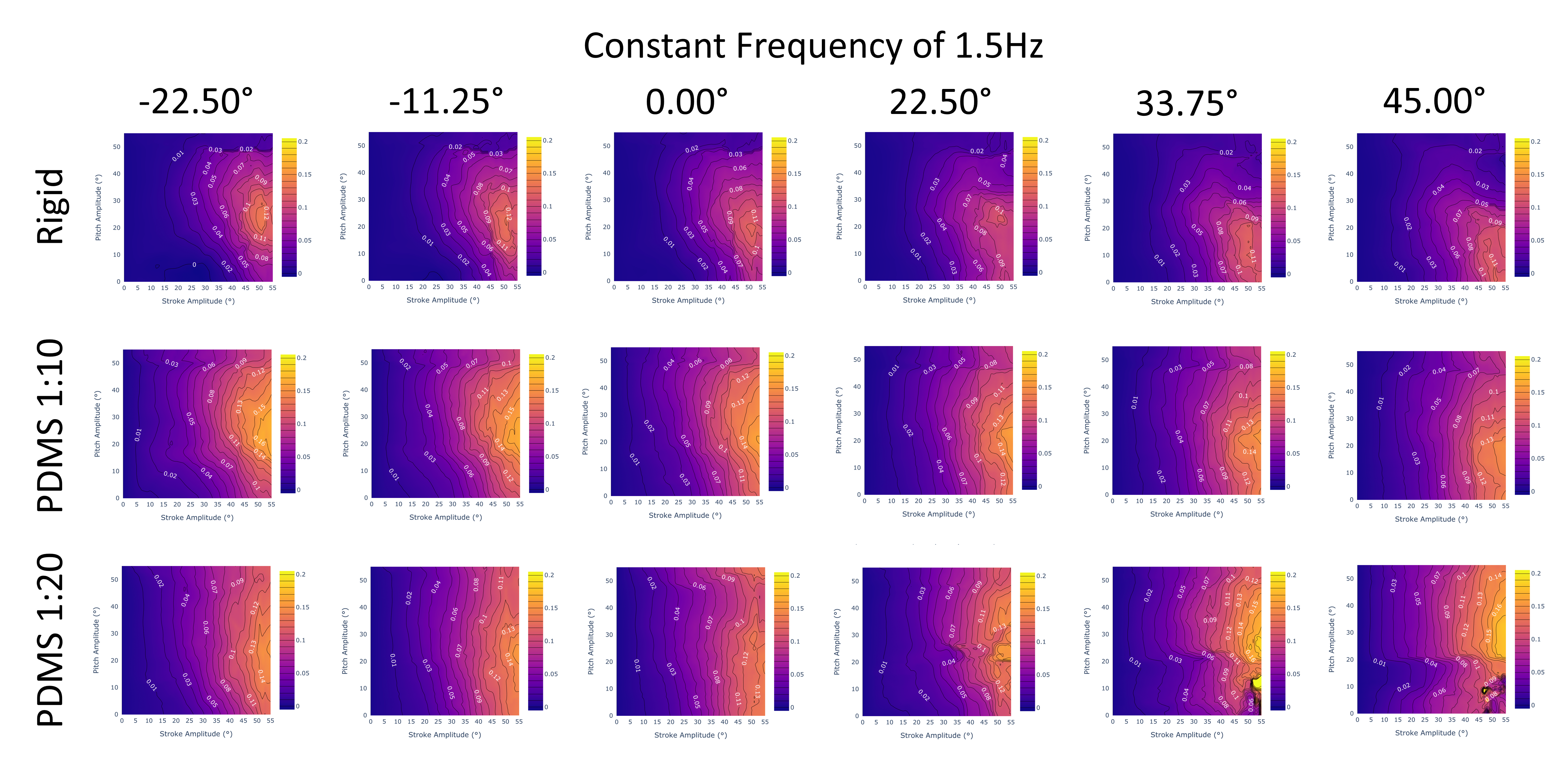}
\caption{Figure of Merit results for a constant frequency of 1.5Hz and varying Stroke-Pitch-Offset. The rows graph all interpolated SPO values with 11.25$^\circ$ increments while the columns are different data sets (rigid, PDMS 1:10, and PDMS 1:20). we verify that the PDMS 1:10 fin is the best performing across all SPO values with higher FOM values and averages compared to the PDMS 1:20 or rigid fins. With the exception of 22.5-45$^\circ$ for PDMS 1:20, a more negative SPO will slightly improve the FOM regardless of fin design.}
\label{fig:spo}
\end{figure*}

This analysis allows us to conclude that the PDMS 1:10 fin design is the most efficient out of all 3 designs, with the highest thrust generation and figure of merit values. Following in second is the PDMS 1:20 fin, which has the second largest thrust generation and figure of merit values. These trends are confirmed across the ranges of stroke and pitch (Figure \ref{fig:fom}) as well as frequency (Figure \ref{fig:freq}) and SPO (Figure \ref{fig:spo}). This suggests that the most efficient design that is able to generate the largest thrust may lie between the two, and is something of interest for future exploration.  

From Figure \ref{fig:freq}, we observe that across the entire range of frequencies, the PDMS 1:10 outperforms both the rigid and PDMS 1:20 fin designs. Additionally, we observe that regardless of fin design, increasing the frequency will improve the FOM metric.

From Figure \ref{fig:spo}, we observe that across the entire range of stroke-pitch offset, the PDMS 1:10 outperforms both the rigid and PDMS 1:20 fin designs. Additionally, we observe that regardless of fin design, a more negative offset will slightly improve the FOM metric, although the difference is very marginal. At high SPO, the PDMS 1:20 fin design appears to diverge from the expected trends at 22.5$^\circ$ or higher and invites future exploration. In the future, revisiting the model's training data for high SPO will likely resolve the issue. 

We conclude the following from the data interpolated: 
\begin{enumerate}[noitemsep]
    \item The best performing fin design is PDMS 1:10, followed by PDMS 1:20. Both consume similar amounts of power, but PDMS 1:10 fins produce a higher thrust. The rigid fin consumes more power and produces less thrust.
    \item The most optimal fin design likely lies in between the PDMS 1:10 and 1:20 fins.
    \item The most efficient gait will occur at a high stroke (40-55), centered pitch (20-35), high frequency (2 Hz) and low SPO (-22.5$^\circ$). 
\end{enumerate}

\section{Inverse Control}
Shown in Figure \ref{fig:inversesystem}, the inverse model communicates with the control system to understand the current state and search for the next gait. The controller feeds the inverse model three metrics: the current gait, the desired thrust, and a set of performance metric weights (for propulsion, kinematics smoothness, and power) that guide the search algorithm. The inverse model samples possible gaits and uses a multi-objective loss function integrated into a Generalized Pattern Search (GPS) method to find a gait that is as close as possible to the desired thrust outcome and meets the priorities weighted. This search algorithm invokes both forward gait-to-thrust and gait-to-power models to accurately predict the thrust and power outcomes of a sampled gait. The finalized gait is sent to the control system. The error signal calculation updates the current position to find the next desired thrust for the inverse model, and another cycle will begin.

\subsection{Objectives}
A common problem with having a live cycle to cycle inverse control model is the speed that models need to search the gait space and return a search result. Because the maximum flapping frequency of any gait is 2Hz, the inverse model is required to generate a gait within 0.5 seconds at maximum, and often practically 0.4 seconds when accounting for time communicating with the control system and calculating positional error. Each search algorithm will invoke both gait-to-thrust and gait-to-power models dozens or hundreds of times in a single search.

Additionally, because of size and power constraints on the control system itself, the model must meet these time constraints on a Raspberry Pi 4 Model B, and forward gait-to-thrust/power models must be shrunk and compressed to meet size and runtime constraints.

\subsection{Loss Function}
Our inverse model compares various sampled gaits in the possible search space with a loss function defined as the aggregate of three separate loss functions for each metric important to our study: $L_t$ as thrust accuracy, $L_k$ as kinematic smoothness, and $L_p$ as power. This loss function is described in Equation \ref{eq:loss1}. 

\begin{equation}
    \label{eq:loss1}
    L=w_t*L_t+w_k*L_k+w_p*L_p
\end{equation}

To prioritize different objectives when selecting an optimal set of fin gaits, $w_t$, $w_k$, and $w_p$ serve as metric weights for each parameter. Weights will always sum to one so that each loss value is weighted proportionally and trades off with the others, as shown in Equation \ref{eq:weights1}

\begin{equation}
    \label{eq:weights1}
    1=w_{T}+w_{K}+w_{P}
\end{equation}

Thrust accuracy loss $L_t$ is computed as the absolute value difference between $T_{target}$ and $T_{pred}$ (Eq. \ref{eq:loss2}), where $T_{target}$ is the target thrust requested by the controller in the current cycle and $T_{pred}$ is the thrust of the gait the inverse model returns based on the specified weights. To calculate $T_{pred}$, the gait-to-thrust neural network is used to compute the expected thrust generation from the gait.

\begin{equation}
    \label{eq:loss2}
    L_t=|T_{target}-T_{pred}|
\end{equation}

The kinematic smoothness loss accounts for the detrimental effect of frequently moving between gaits with highly deviant kinematics between flapping cycles. Transitioning between similar gaits allows the UUV to undergo a smoother motion, promoting system stability. We define a user-selected equivalent step size such that a change of $s_i$ units for kinematic $i$ has the same kinematic smoothness loss to the system as a change of $s_j$ units for kinematic $j$. The kinematic space is normalized by scaling each dimension by its equivalent step size; then, kinematic loss is calculated as the Euclidean distance between the current and proposed gait. Equation \ref{eq:loss3} defines the kinematic smoothness loss function where $n_k$ is the number of kinematics, and $x_i$ and $y_i$ are the values of kinematic $i$ for the current and proposed gait.

\begin{equation}
    \label{eq:loss3}
    L_k=\left(\sum_{i=1}^{n_k}\left( \frac{\left| y_i-x_i \right|}{s_i} \right)^2\right)^\frac{1}{2}
\end{equation}

Unlike the thrust accuracy loss, which has a requested $T_{target}$ from the control system as an input, the power loss is simply $P_{pred}$, which represents the predicted power consumed from the proposed gait of the inverse model. This is helpful for reducing overall power consumption; many gaits can have the same thrust outcome but use different amounts of power based on the combination of kinematics. With $P_{pred}$, any gait which has the same values for $L_t$ and $L_k$ will automatically reward the gait which consumes less power. $P_{pred}$ is calculated using the gait-to-power neural network.

From the Figure of Merit developed in Section 4, it is also possible to calculate a non-dimensional efficiency loss defined in Equation \ref{eq:lossfom}, which is useful for comparing between different materials, shapes, and hardware that may have different thrust outputs and power consumption for the same gait. For example, this is useful to compare between the rigid and flexible designs, as the rigid fin often consumes more power for less thrust with the same set of kinematics.

\begin{equation}
    \label{eq:lossfom}
    L_\eta  = \frac{L_{t}*v}{L_{e}} = \frac{|T_{target}-T_{pred}|*v}{P_{pred}}
\end{equation}

\subsection{Thrust-Kinematics Inverse Control Model}
In previous work by \citet{lee_data-driven_2023}, we demonstrated the effectiveness of a two-weight inverse control model between thrust and kinematics on a rigid fin where the power loss is not considered. We developed and tested various search algorithms where the input space was restricted to attainable gaits described in Equation \ref{eq:1}. The input space is normalized through compressions in the direction of each static kinematic setting $i$ by the corresponding equivalent step size of $s_i$ units; in the normalized input space, any movement between two points of distance $d$ has the same kinematic loss. Three search algorithms including Monte Carlo (MC) \citep{korese14,audet}, Hooke-Jeeves Pattern Search (HJPS) \citep{hooke__1961}, and Generalized Pattern Search (GPS) \citep{doi:10.1137/S1052623493250780,javed_novel_2016, herrera_generalized_2015} algorithms were tested.


All three search algorithms met benchmarking requirements when tested on a Pi 4 Model B, consistently running within 0.5 seconds. While all three models demonstrated a higher accuracy across the weights, the GPS consistently outperformed all three alternatives (Figure \ref{fig:weightsloss}). This is because GPS is able to create multiple searches and exit situations where the algorithm otherwise would be trapped in a local minimum in the gait space with a HJPS algorithm. Additionally, GPS algorithms are a fast derivative-free optimization method that converges fast. GPS operates by searching and polling repeatedly through a mesh of the gait space, where a number of gaits in the mesh are searched and compared to the current gait. When an improvement is found, GPS accepts the gait; when an improvement is not found, GPS will compare mesh points that neighbor the current gait, repeating both steps continuously until convergence.


As such, we proceeded with using the GPS for our inverse control model. A further explanation of all three algorithms can be found in \citep{lee_data-driven_2023}.

\subsection{Inverse Model Testing}
Three separate inverse control models are developed using the characteristics of each material's forward gait-to-thrust and gait-to-power neural networks. All three models meet benchmarking requirements when tested on a Raspberry Pi Model 4B: the rigid inverse control model returns a result after $0.208252 \text{s}$ on average, the PDMS 1:10 inverse control model returns a result after $0.201874 \text{s}$ on average, and the PDMS 1:20 inverse control model returns a result after $0.191625 \text{s}$ on average.

To test the performance gains provided from the inverse control model with three optimization metrics $T$, $K$, and $P$, we run a series of simulations using models of the UUV dynamics MATLAB’s Simulink environment \citep{simulink} to characterize vehicle response to a series of time-varying requested positions (locations the UUV should be at) and velocities (desired speed of the UUV). We conduct a comprehensive test that evaluates the optimal performances for three varying fin materials, including rigid, PDMS 1:10, and PDMS 1:20 material fins. 

\begin{figure}[h!]
    \centering
    \includegraphics[width=0.8\textwidth]{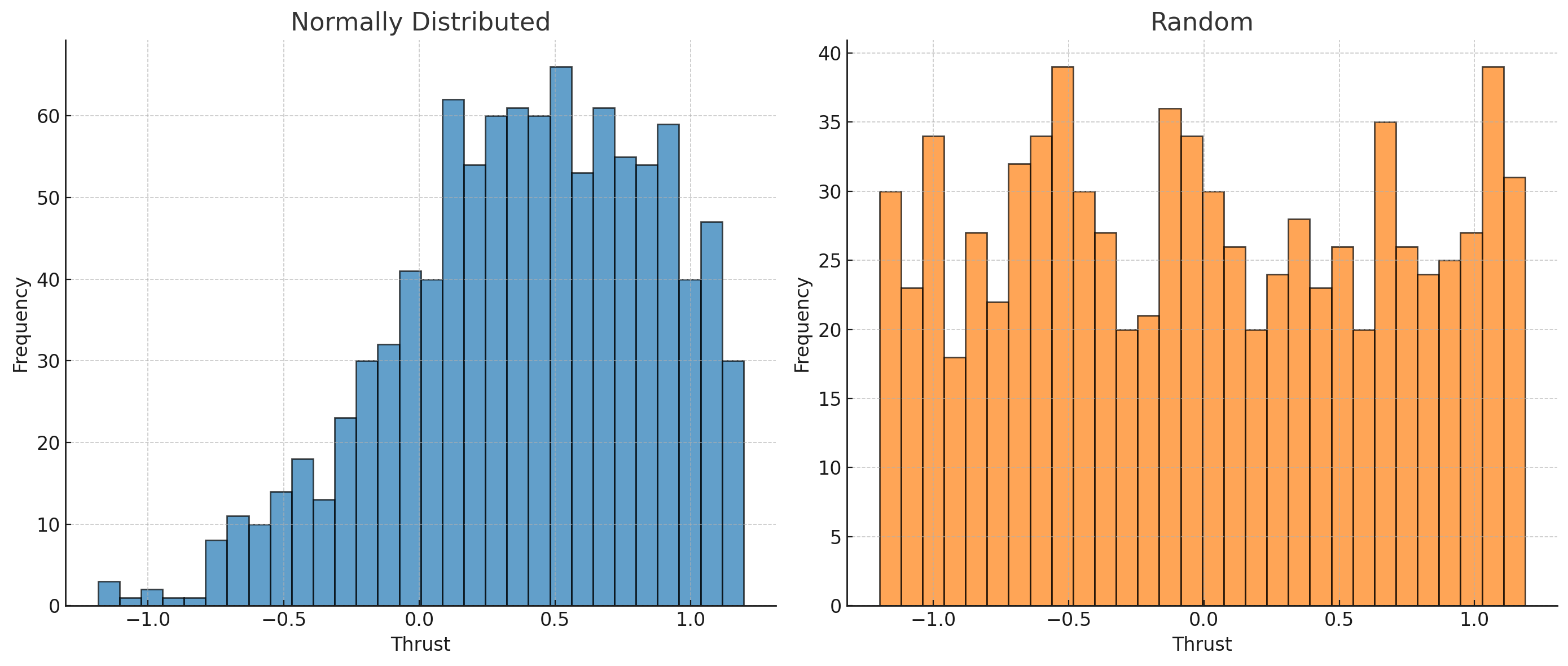}
    \caption{The distribution of the 1010 thrust requests of the random and normal tests.}
    \label{fig:distributions}
\end{figure}

Additionally, a random sample of tests is useful to determine the performance of various combinations of weights to test for the optimal combination of weights to prompt the inverse model for various desired outcomes that are more propulsion-focused or efficient. We create two tests of randomly-generated thrust requests, both containing 1010 tests in total, with distributions shown in Figure \ref{fig:distributions} and summary statistics in Table \ref{tab:summary_statistics}. The first ten gaits will be removed to account for the start from rest in the simulation. 

The first test is fully random from $-1.2N$ to $1.2N$. However, in real-world use, the likelihood that a negative thrust, which is the representation of backwards movement, will be used is unlikely. As such, we create the second test to be normally distributed with a mean of $0.5 N$ and a standard deviation of $0.594 N$, which allocates approximately $20\%$ of values in the distribution to be a negative thrust request. 

\begin{table}[h]
\centering
\begin{tabular}{lrrrrrrrr}
\hline
\textbf{Description} & \textbf{Mean} & \textbf{St. Dev.} & \textbf{Min} & \textbf{Q1} & \textbf{Median} & \textbf{Q3} & \textbf{Max} & \textbf{\% Below 0} \\
\hline
Random & -0.00334 & 0.69586 & -1.19816 & -0.59668 & -0.02966 & 0.61336 & 1.18608 & 51.87\% \\
Normal Dist. & 0.39190 & 0.47204 & -1.18449 & 0.08989 & 0.42847 & 0.75674 & 1.19486 & 20.40\% \\
\hline
\end{tabular}
\caption{Summary statistics for both datasets.}
\label{tab:summary_statistics}
\end{table}


\subsection{Inverse Model Results}

\begin{figure}[h!]
    \centering
    \includegraphics[width=0.8\textwidth]{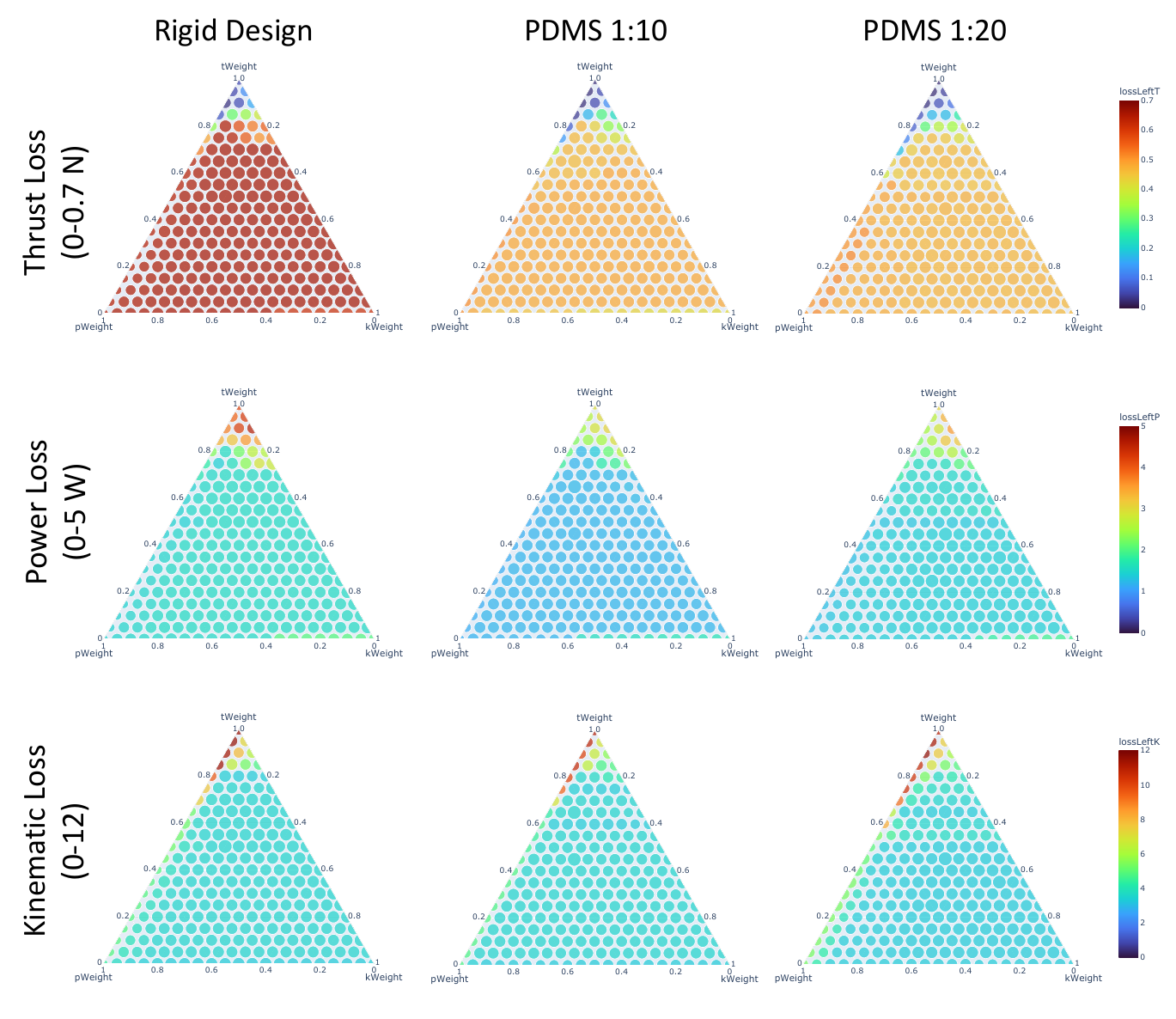}
    \caption{Analysis of different weight combinations. Red indicates high loss, blue indicates minimized loss.}
    \label{fig:allexpsresults}
\end{figure}

Figure \ref{fig:allexpsresults} displays many runs of the test created in the Simulink environment. The standard set of thrust requests is repeated for all possible combinations of optimization weights with a 0.05 difference. Visibly, power loss and kinematics loss exhibit similar behaviors; weighting either against thrust has similar behavior at high values. The most important areas of each triangle are where $w_T>0.7$, where a significant observable trade off between thrust and power/kinematics smoothness exists. This is consistent what we would expect; as the GPS algorithm is basing its search off of a requested thrust from the control system, any algorithm that assigns a low weight to $w_T$ will result in convergence to the same gaits.

When $w_T>0.8$, $L_T$ is generally below 0.2 N for all materials. While improvements can be made to minimize thrust loss to 0.06N as $(w_T, w_K, w_P)$ approaches $(1, 0, 0)$, a generally efficient combination around $(0.8, 0, 0.2)$ can reduce the power consumption of the gaits used by more than two thirds; for a 0.1 N trade off in thrust loss, gait power decreases from a prior 4.2 W to approximately 1.5 W, significantly reducing the power consumption in a total range of 0-1.2 N and 0-7.9 W. These various combinations demonstrate optimal weights for certain tasks, such as rapid acceleration, efficient movement, and energy-efficient station-keeping.

While the profile of each set of optimization weights relative to its neighbors is similar, the flexible PDMS 1:10 and PDMS 1:10 designs have lower thrust losses and lower power requirement compared to the rigid fin. While the rigid fin averages a thrust loss of 0.7 N and a power of 1.5 W, the flexible designs average a thrust loss of around 0.5 N with a power loss of around 1.3W. Similar to previous findings in 4.3, the PDMS 1:10 design is the most effective at minimizing both thrust and power loss simultaneously. 

\begin{figure}[h!]
    \centering
    \includegraphics[width=\textwidth]{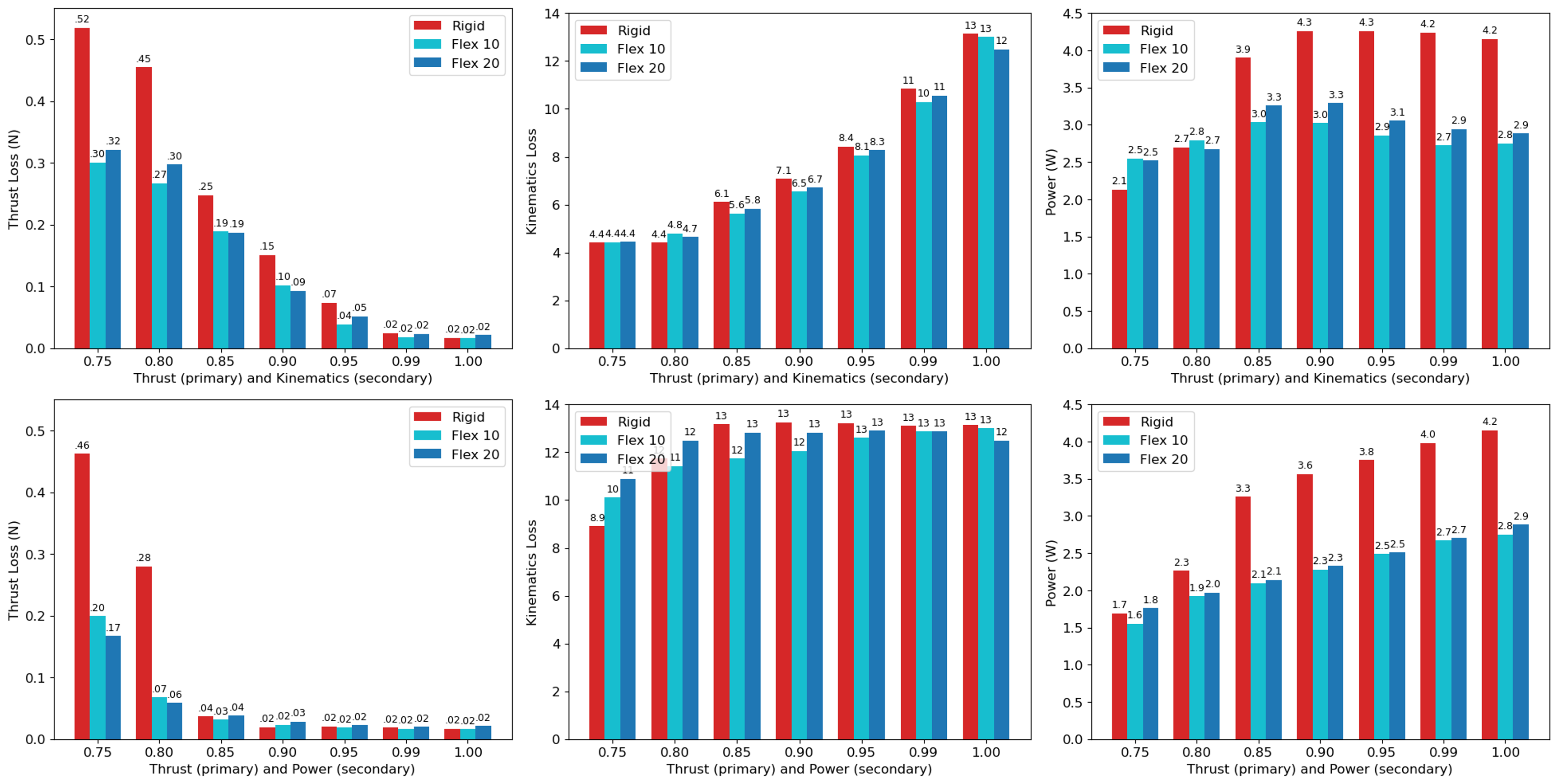}
    \caption{A grid comparing $L_T$, $L_K$, and $L_P$ in each column for various combinations of a high thrust weight trading off with either kinematics (row 1) or power (row 2) on the random test. Inverse models for each fin material are shown side-by-side, with rigid materials in \textcolor{red}{red} and PDMS mixtures in \textcolor{blue}{blue}.}
    \label{fig:randomtesting}
\end{figure}

However, how do we know how any weight will perform if given any gait and thrust request? We can utilize the random and normally distributed datasets (Figure \ref{fig:distributions}) to test how effective an inverse model is in optimizing cycle-by-cycle adjustments to optimize propulsion or efficiency, as shown in Figure \ref{fig:randomtesting}. 

As the simulated gaits span 2.4 N and often switch from flapping forwards to backwards, the kinematics loss is understandably very high, and all three materials perform similarly in optimizing kinematics loss. However, the flexible PDMS materials both outperform the rigid material in minimizing both thrust loss and power. When the weights are $(0.8, 0.0, 0.2)$, the flexible fins cut $L_T$ by more than 75\%, and consistently halves $L_T$ for values where $w_T<0.8$. At values where $w_T>0.85$, the flexible fins improve upon the power loss of the rigid fins by an average of 37\%.

\begin{figure}[h!]
    \centering
    \includegraphics[width=\textwidth]{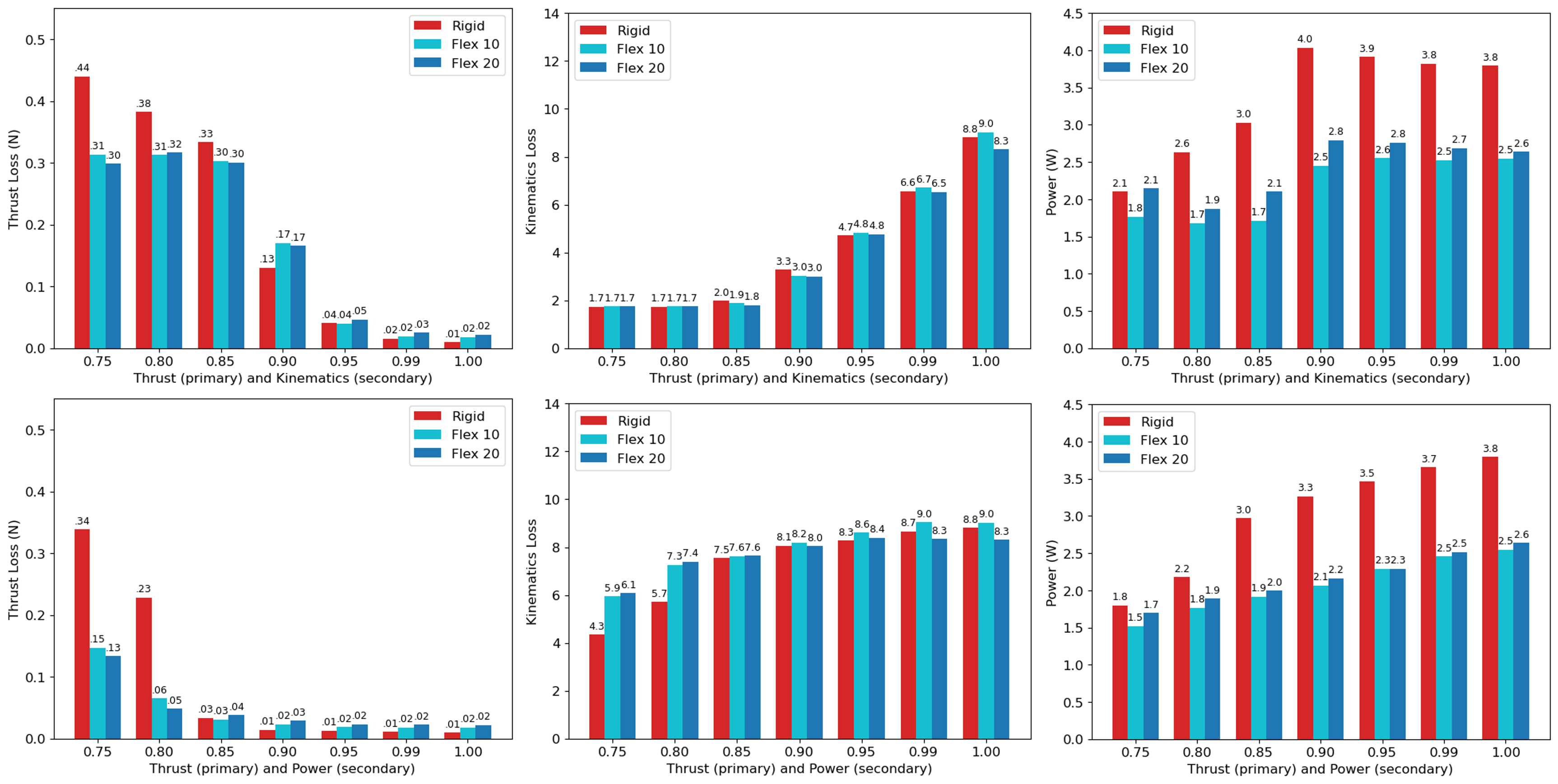}
    \caption{A grid comparing $L_T$, $L_K$, and $L_P$ in each column for various combinations of a high thrust weight trading off with either kinematics (row 1) or power (row 2) on the normally distributed test. Inverse models for each fin material are shown side-by-side, with rigid materials in \textcolor{red}{red} and PDMS mixtures in \textcolor{blue}{blue}.}
    \label{fig:normaltesting}
\end{figure}

The normally distributed test in Figure \ref{fig:normaltesting} confirms prior findings from Figure \ref{fig:randomtesting}. As there is less drastic changes in cycle-to-cycle thrust requests, $L_K$ improves across the board for all three fin materials. Similar behaviors with the random test are observed; the flexible PDMS materials both continue to outperform the rigid material in minimizing both thrust and power loss. At $(0.8, 0.0, 0.2)$, the flexible fins cut $L_T$ by approximately 74\%. Although the flexible fins still reduces $L_T$ for values where $w_T<0.8$, it no longer halves the loss when trading off $w_T$ with $w_K$. The flexible design only reduces $L_T$ by approximately a quarter, but does reduce $L_P$ further. At values where $w_T>0.85$, the flexible fins improve upon the power loss of the rigid fins by an average of 37\%.

\begin{figure}[h!]
    \centering
    \includegraphics[width=\textwidth]{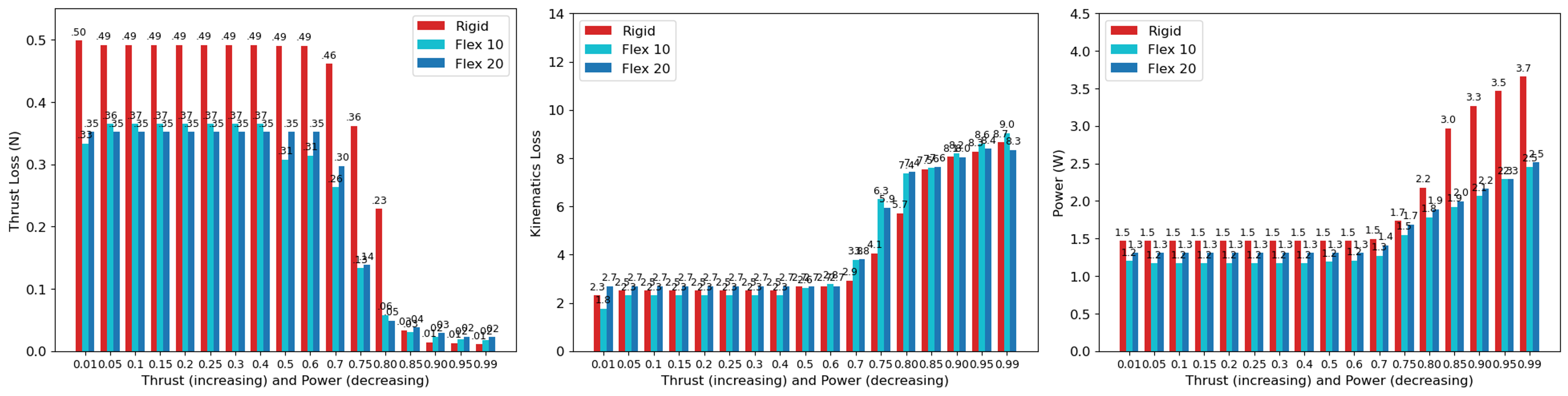}
    \caption{The full range of $w_T$ (increasing) trading off with $w_P$ (decreasing) for each material and loss.}
    \label{fig:weightsloss}
\end{figure}

While these insights are useful, a larger picture that captures the full relationship between $w_T$ trading off with $w_P$ is important. As such, we expand on a larger range that ranges from $(0.99, 0.0, 0.1)$ to $(0.1, 0.0, 0.99)$ for the normally distributed set of thrust requests, shown in Figure \ref{fig:weightsloss}. For the range where $w_T<0.6$, $L_T$ is maximized at 0.49 N. Combined with the average thrust request of 0.39 N, $L_T$ effectively can't go any higher. Figure \ref{fig:weightsloss} ultimately confirms prior findings suggesting that power is strongly correlated with kinematics and shows that thrust and power trade off the most between $0.6<w_T<0.8$ and thus $0.2<w_P<0.4$.




We can also apply the Figure of Merit developed in Section 4 to evaluate the efficiency of the movements at each step. Based on the insights from Figure \ref{fig:weightsloss}, we will only explore the range from 0.75 to 1. We will begin by calculating the average speed of the fin tip throughout the stroke. For the specific fin shape we are testing, the fin span is 15 cm with a rotation axis to the fin root of 3.125 cm, resulting in $r_{tip}=0.18125m$ for Equation \ref{eq:fom4}.

\begin{figure}[h!]
    \centering
    \includegraphics[width=0.6\textwidth]{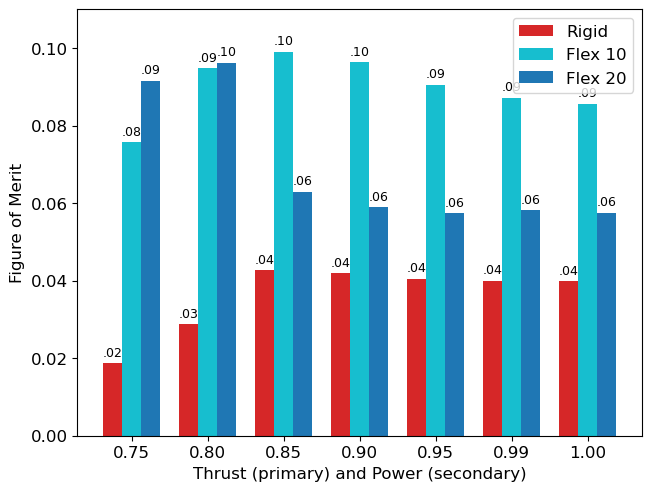}
    \caption{The average FOM of gaits at each weight combination.}
    \label{fig:inversefom}
\end{figure}

This allows us to create Figure \ref{fig:inversefom}, which indicates that the gaits of both flexible fins are very efficient around $(0.8, 0.0, 0.2)$. When $w_T$ increases, both the PDMS 1:10 and 1:20 figure of merits decline. This is expected; the thrust loss is already below 0.01N and will only continue to consume more power. While the PDMS 1:20 fin declines much more due to the flexibility requiring much higher power for the same thrust outcome at higher frequencies, when $w_T<0.8$, the PDMS 1:20 material gains a small performance advantage over the other flexible PDMS 1:10 fin.

\begin{figure}[h!]
    \centering
    \includegraphics[width=0.6\textwidth]{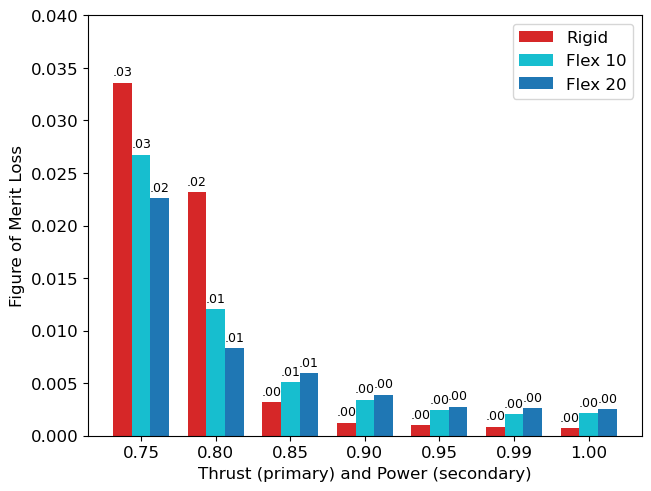}
    \caption{The average non-dimensional thrust loss of gaits at each weight combination.}
    \label{fig:inversefomloss}
\end{figure}

However, the figure of merit of the simulated gaits is not useful when $T_{pred}$ differs from $T_{target}$ substantially. As such, we also calculate the figure of merit 'loss' (Equation \ref{eq:fom6}), which gives an idea of how much dimensionless figure of merit may be deviating. From Figure \ref{fig:inversefomloss}, when $w_T<0.8$, there is a small deviation of $\pm0.035$, negligible for the overall FOM. 

\begin{equation}
    \label{eq:fom6}
    L_{\eta}  = \frac{|T_{target}-T{pred}|*v}{P_{avg}}
\end{equation}

\section{Discussion}
We use two forward-passing LSTM models to generate kinematic interpolations of gait-to-thrust and gait-to-power with the goal of integration onto a control system to optimize for the efficiency of gaits and provide a better understanding of material designs and their relation to efficiency. 

We evaluate six different models for thrust and power to interpolate between experimental data with high accuracy: the linear model, quartic polynomial model, Convolutional Neural Network, Dense Neural Network, and Long-Short-Term Memory model. All six evaluated models accomplish all three criterion we laid out. They are able to:
\begin{itemize}[noitemsep]
    \item Complete a baseline model that inputs gait parameters to output thrust or power with minimal error
    \item Retrain on different designs and maintain a similar or better accuracy
    \item Run at a speed and size suitable for integration onto a control system ($>$100 computations per second)
\end{itemize}
Out of all six models, we find that the LSTM model is able to produce the most accurate results on both the full data set and when we remove specific gaits to create interpolations. We further demonstrate that the LSTM is able to accurately interpolate to unseen and withheld gaits with minimal error.

Using the generated interpolations, we develop a dimensionless Figure of Merit that is able to compare the fin efficiency to other flapping systems and evaluate the efficiency of gaits onboard the control system. 

With the FOM, we conclude that both PDMS materials are more efficient than the rigid fin, with the PDMS 1:10 fin generating the maximum thrust with the lowest power consumption. There were observable trends consistent across all materials, with a higher frequency, stroke amplitude, and negative Stroke-Pitch-Offset all contributing to a greater efficiency. The most efficient gait was concluded to be for the PDMS 1:10 data set with a -22.5$^\circ$ Stroke-Pitch-Offset and frequency of 2 Hz. This understanding will allow us to both design fins that generate a higher thrust and maintain the highest power efficiency, and tune an inverse search model to search gaits it knows to be power-efficient or optimal. 

We implemented inverse models incorporating both the gait-to-thrust and gait-to-power forward models within the time constraint of 0.5s per iteration with the onboard hardware. Incorporating the gait-to-power model reduced the time to search for an answer for all three inverse models. Each model is able to make trade-offs and optimize between thrust accuracy, kinematics smoothness, and power consumption based on current cycle-to-cycle needs. The flexible inverse model framework enables future UUV control systems to incrementally adjust the emphasis placed on different measures of performance based on the current task and vehicle status. For example, the emphasis on thrust accuracy can be dynamically changed by the controller based on the degree of precise maneuvering required for the task at hand. The inverse model framework also allows for the incorporation of additional performance metrics such as efficiency.

Using our three inverse models, we are able to search for optimal combinations that improve general efficiency. Our inverse model validates our figure of merit's findings: flexible fins, especially the PDMS 1:10 mixture, can improve thrust outcomes by upwards of 75\% and power outcomes upwards of 66\% for the same set of thrust requests based on the specific weights. Throughout simulation and analysis, $(0.8, 0.0, 0.2)$ emerged as a point where thrust often trades off with either kinematics smoothness or power, drastically increasing or decreasing depending on the direction. However, this weight set serves as a base-level optimization that improves thrust outcomes by 29\% and power outcomes by approximately 33\%. Kinematics smoothness remains similar between all three materials. 

We are currently in the deployment stage, collecting experimental data for in-tank testing of the control system to evaluate and, if needed, retrain the forward model by running physical simulations in underwater environments. Inverse model performance for thrust accuracy and propulsive efficiency trade-offs will be evaluated. The results of full simulated trials, including insights into optimal weight combinations for varying fin flexibilities and developed weight combinations tailored for various operational requirements (i.e. station keeping or rapid acceleration) will be further detailed, with recommendations made to set ‘default’ settings for certain tasks or missions. These will accommodate varying fin materials and test settings. 

\section{Conclusion}
This study creates material insights for the effect of flexibility on propulsion and efficiency outcomes, while using deep learning to optimize the propulsion and power consumption of the same movement by searching for and selecting a suitable gait. We develop and test an inverse search model that utilizes a forward gait-to-thrust and forward gait-to-power to predict the thrust and power outcomes of various kinematics using a LSTM model with an accuracy of 0.0076 N and 0.0072 W. This allows our inverse search model, benchmarked within necessary time constraints, to prioritize between thrust propulsion, kinematics smoothness, and power use for each individual gait. By default, use of the inverse model reduces the thrust loss and power consumed for any requested thrust by an average of 0.5 N and 3.0 W. By using a dimensionless figure of merit, we can conclude that both flexible PDMS 1:10 and 1:20 materials are more efficient and propulsive compared to the rigid fin, with the PDMS 1:10 fin being the most efficient and propulsive overall; with the same gait, the 1:10 mixture improved thrust outcomes by upwards of 75\% and power outcomes by upwards of 66\%. Improvements from inverse search are amplified with the 1:10 material. For any material, inverse search methods improve thrust outcome by 29\% and reduce power consumption by 33\% simultaneously; weights that prioritize either can further optimize for either propulsion or efficiency. 


\newpage
\bibliographystyle{apacite}
\bibliography{references} 

\end{document}